%% file: main.tex
\documentclass[acmsmall]{acmart}

\usepackage{graphicx}
\usepackage{float}
\usepackage{booktabs}
\usepackage{caption}
\usepackage{geometry}
\usepackage{flushend}
\usepackage{float}
\usepackage{subfigure}
\AtBeginDocument{%
  \providecommand\BibTeX{{%
    \normalfont B\kern-0.5em{\scshape i\kern-0.25em b}\kern-0.8em\TeX}}}
\usepackage{tabularx}
\usepackage{bbding}
\usepackage{graphicx}
\usepackage{caption}
\usepackage{geometry}
\usepackage{amsmath}
\usepackage{makecell}
\usepackage{float}
\usepackage{color}
\usepackage{booktabs}
\usepackage[normalem]{ulem}
\usepackage{tabu}
\usepackage{hyperref}
\usepackage{xcolor}
\usepackage{appendix}

\usepackage{xcolor}   % 用于颜色
\usepackage{quoting}  % 用于引言环境

\newenvironment{dialogue}
  {\begin{quote}\ttfamily}
  {\end{quote}}

\setcopyright{acmlicensed}
\copyrightyear{2018}
\acmYear{2018}
\acmDOI{XXXXXXX.XXXXXXX}

\acmConference[Conference acronym 'XX]{Make sure to enter the correct
  conference title from your rights confirmation emai}{June 03--05,
  2018}{Woodstock, NY}
\acmISBN{978-1-4503-XXXX-X/18/06}

\author{Ruoyu Wen}
\email{ruoyu.wen@pg.canterbury.ac.nz}
\orcid{0009-0008-0052-0045}
\authornotemark[1]
\affiliation{%
  \institution{University of Canterbury}
  \streetaddress{20 Kirkwood Avenue, Upper Riccarton}
  \city{Christchurch}
  \state{Canterbury}
  \country{New Zealand}
  \postcode{8041}
}

\author{Stephanie Elena Crowe}
\email{stephanie.crowe@pg.canterbury.ac.nz}
\orcid{0000-0002-6821-1993}
\authornotemark[2]
\affiliation{%
  \institution{University of Canterbury}
  \streetaddress{20 Kirkwood Avenue, Upper Riccarton}
  \city{Christchurch}
  \state{Canterbury}
  \country{New Zealand}
  \postcode{8041}
}

\author{Kunal Gupta}
\email{kunal.gupta@auckland.ac.nz}
\orcid{0000-0003-3963-8856}
\authornotemark[2]
\affiliation{%
  \institution{University of Auckland}
  \streetaddress{70, Symonds Street}
  \city{Auckland}
  \state{Auckland}
  \country{New Zealand}
  \postcode{1010}
}

\author{Xinyue Li}
\email{lixinyue_e@163.com}
\orcid{0009-0002-8177-9775}
\authornotemark[3]
\affiliation{%
  \institution{Uppsala University}
  \country{Sweden}  
}

\author{Mark Billinghurst}
\email{mark.billinghurst@auckland.ac.nz}
\orcid{0000-0003-4172-6759}
\affiliation{%
  \institution{The University of Auckland}
  \city{Auckland}
  \state{Auckland}
  \country{New Zealand}
  \postcode{1010}
}

\author{Simon Hoermann}
\email{simon.hoermann@canterbury.ac.nz}
\orcid{0000-0002-4201-844X}
\affiliation{%
  \institution{University of Canterbury}
  \streetaddress{20 Kirkwood Avenue, Upper Riccarton}
  \city{Christchurch}
  \state{Canterbury}
  \country{New Zealand}
  \postcode{8041}
}

\author{Dwain Allan}
\email{dwain.allan@canterbury.ac.nz}
\orcid{0000-0001-9859-7033}
\affiliation{%
  \institution{University of Canterbury}
  \streetaddress{20 Kirkwood Avenue, Upper Riccarton}
  \city{Christchurch}
  \state{Canterbury}
  \country{New Zealand}
  \postcode{8041}
}

\author{Alaeddin Nassani}
\email{alaeddin@u-aizu.ac.jp}
\orcid{0000-0002-3587-5523}
\affiliation{%
  \institution{University of Aizu}
  \city{Fukushima}
  \state{Fukushima}
  \country{Japan}
}

\author{Thammathip Piumsomboon}
\email{tham.piumsomboon@canterbury.ac.nz}
\orcid{0000-0001-5870-9221}
\affiliation{%
  \institution{University of Canterbury}
  \streetaddress{20 Kirkwood Avenue, Upper Riccarton}
  \city{Christchurch}
  \state{Canterbury}
  \country{New Zealand}
  \postcode{8041}
}

\begin{document}
%%\citestyle{authoryear}
%\title[LLMs for Automatic Detection of Sensitive Topics]{Evaluating Large Language Models for Automatic Detection of Sensitive Topics in Online Counselling}
\title[LLMs for Automatic Detection of Sensitive Topics]{Large Language Models for Automatic Detection of Sensitive Topics}
% \title[LLM-Detector]{Automatic Detection of Sensitive Topics in Online Counseling Based on Large Language Models}

\begin{abstract}
Sensitive information detection is crucial in content moderation to maintain safe online communities. Assisting in this traditionally manual process could relieve human moderators from overwhelming and tedious tasks, allowing them to focus solely on flagged content that may pose potential risks. Rapidly advancing large language models (LLMs) are known for their capability to understand and process natural language and so present a potential solution to support this process. This study explores the capabilities of five LLMs for detecting sensitive messages in the mental well-being domain within two online datasets and assesses their performance in terms of accuracy, precision, recall, F1 scores, and consistency. Our findings indicate that LLMs have the potential to be integrated into the moderation workflow as a convenient and precise detection tool. The best-performing model, GPT-4o, achieved an average accuracy of 99.5\% and an F1-score of 0.99. We discuss the advantages and potential challenges of using LLMs in the moderation workflow and suggest that future research should address the ethical considerations of utilising this technology.
\end{abstract} 

%%
%% The code below is copied from, generated by the tool at http://dl.acm.org/ccs.cfm.
\begin{CCSXML}
<ccs2012>
   <concept>
       <concept_id>10003120.10003130.10011762</concept_id>
       <concept_desc>Human-centered computing~Empirical studies in collaborative and social computing</concept_desc>
       <concept_significance>500</concept_significance>
       </concept>
 </ccs2012>
\end{CCSXML}

\ccsdesc[500]{Human-centered computing~Collaborative and social computing~Collaborative and social computing systems and tools}

%%
%% Keywords.
\keywords{Online Moderation, Automatic Detection, Large Language Model, Comparative Study}

% \centering
% \subfigure[]{
% \includegraphics[width=0.305\textwidth, trim= 0 0 40 3]{figs/cover1.JPG}\label{fig1a}
% }\hspace{1mm}
% \subfigure[]{
% \includegraphics[width=0.305\textwidth, trim= 0 0 50 0]{figs/cover2.JPG}\label{fig1b}
% }\hspace{1mm}
% \subfigure[]{
% \includegraphics[width=0.315\textwidth, trim= 0 0 50 0]{figs/cover3.JPG}
% \label{fig:01}
% }
% \caption{Caption}
% \Description{Caption}


\maketitle
\vspace{-0.36cm}

\section{Introduction}\label{sec:Introduction}
\input{sections/01-Intro.tex}

\section{Related Work}\label{sec:Background}
\input{sections/02-Back.tex}

\section{Methodology}\label{sec:Methodology}
\input{sections/03-Methodology}

\section{Preliminary Study}\label{sec:Preliminary}
\input{sections/04-Pre}

\section{Formal Study}\label{sec:Formal}
\input{sections/05-Formal}

\section{Formal Study Results}\label{sec:Results}
\input{sections/06-Results}

\section{Discussion}\label{sec:Discussion}
\input{sections/07-Discussion}

\section{Conclusion}\label{sec:Conclusion}
\input{sections/08-Conclusion}

% \input{sections/Appendix}

% \begin{acks}
% thanks.
% \end{acks}

\bibliographystyle{ACM-Reference-Format}
\bibliography{reference}

\newpage
\newpage
\newpage
\appendix

\label{sec:Appendix}
\input{sections/Appendix.tex}

\end{document}

%% file: sections/01-Intro.tex
This paper explores how Large Language Models (LLMs) can be utilised to identify sensitive information related to mental well-being in the context of human communications. The recent development of LLMs has led to their integration into many traditional workflows. These models are recognised for their capability to comprehend and process natural languages effectively. Given their advanced understanding and generation capabilities, LLMs offer a potential approach to automating the detection of sensitive topics such as suicidal ideation or self-harm, grief and loss, sexual assault, and abuse. This paper evaluates the capability of LLMs to detect such topics. By examining the accuracy and consistency of different LLMs, this study aims to provide insights into their potential to identify and process sensitive information effectively.

One application area for this is content detection and moderation in mental health and the use of conversational agents and chatbots to provide mental health support safely and ethically. The prevalence of mental health issues is escalating globally, presenting a substantial public health challenge. The World Health Organisation (WHO) estimates that one in four people will encounter mental health difficulties at some point in their lives \cite{world2017depression}. Youth (ages 12–24) is identified as a critical period, with the majority of mental disorders commencing during these formative years \cite{patel2007mental}. The protracted COVID-19 pandemic appears to further exacerbate these trends \cite{siegert2023depression,vacchiano2023first}. However, globally, healthcare systems face a pronounced shortage of trained mental health professionals, making it a significant challenge for young people to access timely mental support \cite{kakuma2011human,bruckner2011mental}.

%In New Zealand, there has been a significant rise in the mental health needs of adolescents since 2012, observed across all demographic groups, with a particularly acute increase among female, Māori, Pacific, and Asian students, as well as those residing in high-deprivation neighbourhoods \cite{sutcliffe2023rapid}. 

Mental health chatbots are increasingly recognised as a promising and impactful tool for mental health professionals within the context of mental health support \cite{d2020ai}. Recent studies have demonstrated their ability to assist in managing mental health issues \cite{WoeBotWebsite,CassAIWebsite,WysaWebsite,abd2021perceptions}. There are a number of advantages to using them; Chatbots can be accessed 24 hours a day in multiple languages on various health topics, reducing labour costs and improving counselling efficiency. However, prior studies have indicated that chatbots have limitations in ensuring safety within the mental well-being domain \cite{d2020ai, BBCarticle, ma2023understanding}. One disadvantage is that chatbots often lack the ability to process sensitive information, which can sometimes harm the user's mental well-being instead of supporting it. Consequently, human-in-the-loop monitoring is still crucial in human-AI interactions in real time.

Chatbots can show a progressive ability to understand the context in conversations \cite{zhu2024can}, indicating a potential to support humans with sensitive topic detection. LLMs can help detect sensitive information by labelling sensitive messages and notifying a human moderator to intervene. This study aims to investigate the capabilities of LLMs to detect sensitive topics within the mental well-being domain. Our aim is to test the performance of various LLMs on metrics such as accuracy, precision, and recall in automatically detecting sensitive messages. By conducting this study, we seek to answer two research questions:
\begin{enumerate}
    \item[RQ1:] What is the impact of different LLMs on detecting sensitive topics?
    \item[RQ2:] How does the temperature parameter setting affect LLMs' performance in detecting sensitive topics?
\end{enumerate}

The rest of the paper is organised as follows. Section 2 introduces related work in this field, and Section 3 explains the methodology we followed. While Section 4 introduces the preliminary study, Section 5 covers the procedure of the formal and the results are presented in Section 6. Finally, discussions and conclusions are presented in Sections 7 and 8, respectively. The main contributions of this paper are: 

\begin{enumerate}
    \item We provide one of the first evaluations of the capability of several common LLMs at different temperature settings in detecting mentally sensitive messages.
    \item We curated datasets to form a collection of sensitive and non-sensitive messages.
    \item We provided a framework for testing the capability of LLMs in detecting sensitive content during the moderation process.
\end{enumerate}

%% file: sections/02-Back.tex
Automated content detection is a critical area of research, particularly for identifying sensitive and harmful information on online platforms. This section explores the various methods and technologies employed in this domain, highlighting the evolution from traditional human moderation to advanced machine-learning approaches. Subsection~\ref{subsec:automated-content-detection} explores the application of automated content detection in different contexts, such as preventing cyberbullying and supporting mental health. Subsection~\ref{subsec:perceptions-detection} discusses the perceptions and implications of technology-assisted detection, emphasising the balance between automated tools and human oversight. Finally, Subsection~\ref{subsec:llm-method} introduces LLMs as a new method, showcasing their potential to revolutionise automated content detection with their powerful text generation and understanding capabilities.

\subsection{Automated Content Detection}
\label{subsec:automated-content-detection}
Sensitive and harmful information detection is always an important topic in many areas. Many online forums and social media platforms traditionally employ human moderators to review and manage content manually. This approach is often inefficient. To improve efficiency, some crowd-sourced workers have been introduced in this field \cite{hettiachchi2019towards,bozarth2023wisdom}, but this approach, at the same time, has increased costs. Recently, many studies have explored using automated content detection, often machine-learning-based, to assist human moderators, aiming to replace manual review methods with higher efficiency and lower cost. 

Automated content detection has been extensively researched in the context of preventing cyberbullying and cybercrime \cite{slonje2013nature}. For example, Al-Garadi et al. \cite{al2016cybercrime} developed a machine learning model to detect cyberbullying on Twitter, analysing network, activity, user, and content features from 2.5 million geo-tagged tweets. They achieved high detection accuracy and precision using classifiers like Random Forest, proposing a tool for monitoring and mitigating online bullying. Similarly, Zhao et al. \cite{zhao2016automatic} proposed a new model that enhances traditional text classification by integrating bullying-specific features derived from word embeddings. Their model showed superior performance compared to baseline methods. Dadvar and De Jong \cite{dadvar2012cyberbullying} integrate user characteristics and behaviours across social networks into a new model to detect cyberbullying. Their study offers a tool for more accurate cyberbullying detection and enhanced support for victims and moderators. 

One important domain for identifying sensitive information is E-health, which is a sensitive area that requires careful handling. Huh, et al. \cite{huh2013text} developed a text classification system to assist moderators in identifying posts that need their expertise in online health communities. They used a binary Naive Bayes classifier to categorise posts into those needing a moderator's response and those that do not. This approach provides a low-cost, scalable solution to help moderators efficiently prioritise posts requiring their attention, thereby enhancing support in large-scale health forums. 

There are also some studies about suicidality detection. O'Dea et al. \cite{o2015detecting} developed a machine-learning model to detect suicidality on Twitter. They collected tweets containing suicide-related phrases and used human coders to classify the level of concern. These tweets were then used to train a classifier. The classifier achieved an overall accuracy of 76\%, correctly identifying 80\% of ``strongly concerning'' tweets. Cohan et al. \cite{cohan2017triaging} also used machine learning to build a system to automatically triage the severity of posts in online mental health forums. The system categorises posts into four levels of severity, from no risk to imminent risk of self-harm, using a feature-rich classification framework that includes lexical, psycholinguistic, contextual, and topic modelling features. The system significantly improved the identification of high-risk posts, achieving up to a 17\% improvement in F1 scores, an indicator of precision and recall capability, compared to previous methods. 

Researchers also looked at peer support, which is beneficial for young people to overcome barriers in help-seeking behaviour. Milne et al. \cite{milne2019improving} developed a machine learning system to triage posts in online peer support forums, categorising messages by urgency.  Using a classifier, it achieved 84\% accuracy, improving moderator efficiency and response times on online forums. Milne et al. \cite{milne2016clpsych} organised the ``CLPsych 2016 Shared Task'' to develop systems for automatically triaging posts in online mental health forums, specifically ``ReachOut.com''. Participants classified posts into four categories: Green (no immediate action needed), Amber (moderate concern), Red (high concern), and Crisis (immediate action needed). Using various machine learning techniques, the best-performing systems significantly outperformed baselines.

\subsection{Perceptions of Technology-Assisted Detection}
\label{subsec:perceptions-detection}
From prior studies, we found that machine learning is the most common method for building AI-supported detectors or moderators. It can achieve high accuracy and reduce detection time. Some have already been integrated into practical workflows. Jhaver et al. \cite{jhaver2019human} conducted a study of Automoderator (Automod) on the Reddit platform. They interviewed 16 Reddit moderators, including the creator of Automod, to explore the benefits and challenges associated with automated content detection technologies. The result indicated that Automod significantly reduces the workload of human moderators by handling menial tasks and protecting them from emotional labour. However, using Automod also creates new problems for human moderators. They must check for incorrect decisions, as Automod lacks contextual sensitivity and struggles with cultural and linguistic nuances. Based on the forum regulations, the tool requires regular updates and fine-tuning, adding to the workload of already limited skilled moderators. Additionally, moderators may find it difficult to fully understand and control Automod's actions, especially when it makes mistakes or its decisions are unclear, it's not fully transparent. At last, the paper highlighted the need for a balanced approach that combines automated and human moderation to maintain effective content regulation. 

Many scientists have raised concerns about using technology-assisted tools for content detection \cite{udupa2023ethical,siapera2022ai,nahmias2021oversight}. Gillespie \cite{gillespie2020content} mentioned the inability of these tools to grasp context and nuance, leading to errors in detecting sarcasm and cultural subtleties. Additionally, Gillespie is concerned about the risk of AI perpetuating biases from training data, potentially impacting marginalised groups unfairly. He emphasised the need for human oversight in moderation to handle complexities and ethical considerations, arguing that AI should assist rather than replace human moderators. He pointed out that ``\textit{the kind of judgment that includes the power to ban a user from a platform should only be made by humans.}" Gorwa et al. \cite{gorwa2020algorithmic} examined the technical and political challenges of algorithmic content moderation on major platforms like Facebook, YouTube, and Twitter. Their study underscored the opaqueness and accountability issues of automated moderation and similarly emphasised the need for human oversight to address the complex and contextual nature of content moderation effectively.

\subsection{Large Language Model: A New Method}
\label{subsec:llm-method}
The rapid development of LLMs introduces a new method in this domain, changing the way humans interact with machines \cite{abbilaashat2023chatgpt}. LLMs possess powerful text-generating and understanding capabilities \cite{wu2023brief}. Users can ask AI to explain each decision in natural language simply by using natural language prompts, making communication straightforward. In such cases, integrating LLMs to replace traditional machine learning methods becomes a viable choice for automated content detection.

Wang et al. \cite{wang2023not} evaluated six popular LLMs for detecting and handling harmful instructions. LLaMA-2 showed the best safety performance, while ChatGLM2 was the least effective. Commercial models like GPT-4 often rejected harmful instructions directly. He et al. \cite{he2024if} compared GPT-4's data labelling with 415 crowd-sourced workers from the MTurk platform on 3,177 sentence segments from 200 scholarly papers. GPT-4 achieved 83.6\% accuracy, while MTurk reached 81.5\%. Combining their labels improved accuracy to 87.5\%. The study shows that integrating human and AI labelling enhances accuracy, highlighting the value of combining crowdsourcing with advanced AI models. Kolla et al. \cite{kolla2024llm} evaluated GPT-3.5's effectiveness in moderating Reddit content, analysing 744 posts across nine subreddits. They found GPT-3.5 had a high true-negative rate (92.3\%) but a low true-positive rate (43.1\%), indicating it struggled to accurately flag rule violations. While it handled keyword-based violations well, it missed more complex cases. The study suggests a hybrid approach, combining GPT-3.5 with human moderators, is necessary for effective content moderation. This research investigates five models in our formal study, including GPT-4, GPT-4o, GPT-4-Turbo, Llama-3, and Solar. 

%% file: sections/03-Methodology.tex
\begin{figure} 
    \centering %居中
    \includegraphics[width =1\textwidth]{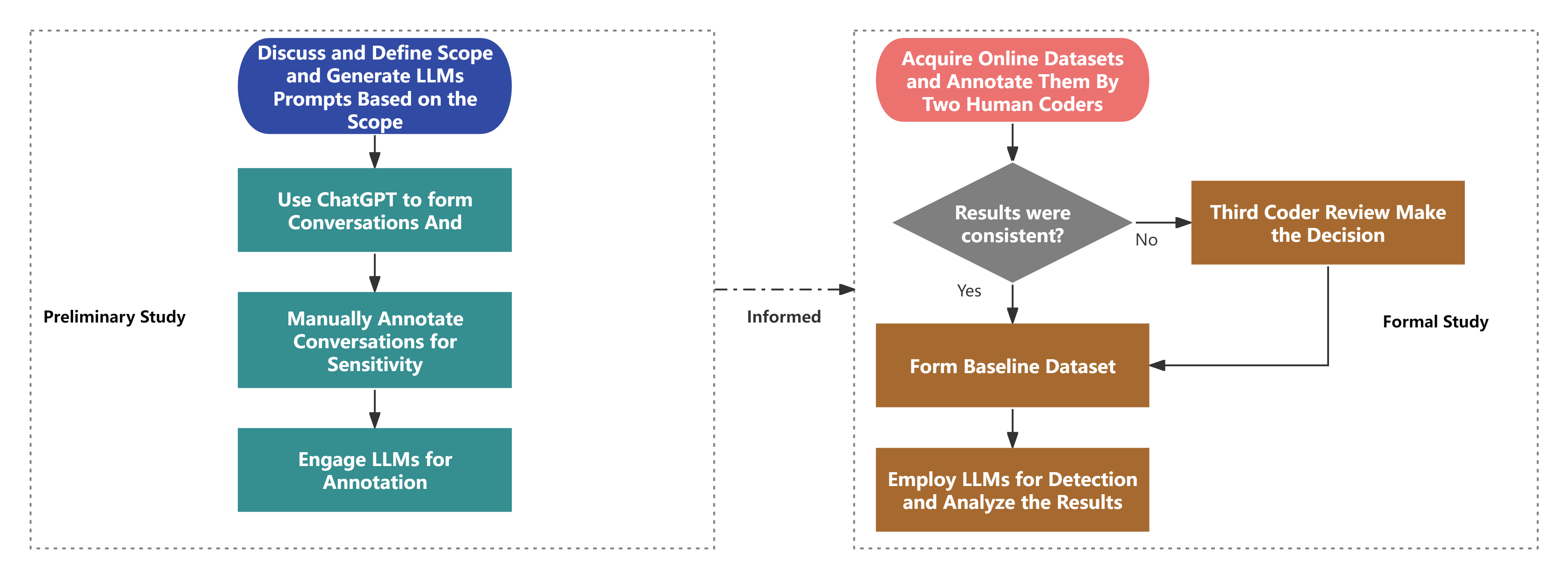} %图片的路径，默认以.tex 所在的文件夹作为前缀
    \caption{The research procedure consists of two phases: the preliminary study and the formal study.} %图片标题
    \label{fig:flow} %图片标签，用于交叉引用
\end{figure}

This research comprises two phases, as shown in \ref{fig:flow}. First, we conducted a preliminary study, detailed in Section~\ref{sec:Preliminary}. This involved using a small dataset to evaluate various LLMs for detecting sensitive topics in conversations among university freshmen. This initial study aimed to gain a general understanding of LLM performance in detecting sensitive topics. It covers dataset preparation, LLM selection and fine-tuning, and data processing. Preliminary results highlight the effectiveness of LLMs, with some models demonstrating high accuracy in identifying sensitive information.

Second, in Section~\ref{sec:Formal}, we conducted a formal study by evaluating five LLMs on a task involving datasets with both sensitive and non-sensitive messages. The procedures for both studies are shown in Fig. \ref{fig:flow}. Section~\ref{sec:Results} reports on the accuracy, precision, recall, F1 scores, and consistency of the models. Additionally, we tested the impact of the temperature parameter on LLM performance and analysed the causes of errors encountered during the study. Finally, Section~\ref{sec:Discussion} discusses the study's results.

%% file: sections/04-Pre.tex
To test the capability of models for the subsequent experimental design, we conducted a preliminary study. In this study, several LLMs were tested to detect whether the topics of conversation were sensitive. The context was set as conversations among university freshmen undergoing a transition. The scope of sensitive topics was based on prior studies \cite{transition,delamar2012supporting} and discussions among the research team. These topics included: `Strong Emotion or Stress,' `Suicidal Ideation or Self-Harm,' `Sexual Assault and Abuse,' `Severe Mental Health Disorders,' `Substance Abuse,' `Eating Disorders,' `Trauma and PTSD,' `Grief and Loss,' `Family Issues,' `Identity and Discrimination Issues,' `Legal Issues and Crime,' `Physical Hurt' and `General Stress and Bad Mood.'

\subsection{Procedure}

\subsubsection{Dataset Preparation}
We used ChatGPT (GPT-4) to generate casual conversations between a university freshman and her best friend about her first-year life in school. the prompt was: 

``\textit{I would like you to play as a freshman named Lily, communicating with your friend Sam about the problems you encounter in your first-year uni life. Lily is a freshman facing transition issues. I don't want continuous dialogue, but ten discontinuous pieces of conversation. I want this communication to start from the beginning like you might say "Hi, how is it going?" (Please use the format: Lily: `...', Sam: `....'.) Some of these are about Lily's troubles with stress in school, and others are casual talk. Generate 500 messages, I want half of the conversations to be mentally safe, and half to be unsafe.}" 

From this generation, we obtained 500 messages(50\% sensitive) to create the pool. Then, a research team member manually checked messages and selected 100 sensitive messages along with 100 non-sensitive messages to form the dataset. Here are some examples of sensitive messages:
\begin{enumerate}
\item \begin{dialogue}
\textbf{Message:}I think I need to talk to someone professional. I'm overwhelmed by my emotions and it's getting harder to cope.

\textbf{Reason: }General Stress and Bad Mood
\end{dialogue}

\item \begin{dialogue}
\textbf{Message:}I know it’s bad, but I feel like I can’t stop. It’s the only thing that helps me feel less anxious.

\textbf{Reason: }Substance Abuse
\end{dialogue}
\end{enumerate}
\subsubsection{LLM Preparation}
We selected five popular and powerful LLMs available at the time of this research for testing, including GPT-3.5-Turbo-0125, GPT-4-Turbo Claude-3-Opus-20240229, Llama-3-70B-8192, and GPT-3.5-Turbo-0125 (Fine-Tuned). Fine-tuning is a process of pre-training an LLM for specific tasks, believed to enhance its performance in specific areas \cite{vm2024fine}. In this study, we choose GPT-3.5-Turbo-0125, the most advanced GPT model for fine-tuning at the time of this research. We selected 70 messages as the training data and 30 as the validation data suggested by OpenAI \cite{openai_2023_ft}. Then, we used the OpenAI Playground \footnote{https://platform.openai.com/playground} to train the fine-tuned model.

\subsubsection{LLM Data Processing}
We used JSON mode in Python 3.9.13 to call the API of each of the five models to batch-process the datasets. The processed data was then stored in CSV files.  The following prompt was given to the LLMs as instructions: 

\textit{``You are a detector that determines whether the input falls within a given range. The range includes Strong Emotion or Stress, Suicidal Ideation or Self-Harm, Sexual Assault and Abuse, Severe Mental Health Disorders, Substance Abuse, Eating Disorders, Trauma and PTSD, Grief and Loss, Family Issues, Identity and Discrimination Issues, Legal Issues, Crime. And something that can cause stress and bad mood, Long sentences of more than 30 words, and Physical hurt or related topics. Simply respond with 'yes' if it is related, or 'no' if it is not.''} 

We set the prompt slightly broader than our scope, assuming that LLMs might struggle with long, complex sentences \cite{jozefowicz2016exploring}.

\subsection{Preliminary Results}
The results are shown in Table~\ref{tab:preliminary}. True Positives (T+) represent the number of correctly identified sensitive data points. True Negatives (T-) represent the number of correctly identified non-sensitive data points. False Positives (F+) represent the number of non-sensitive data points incorrectly identified as sensitive. False Negatives (F-) represent the number of sensitive data points incorrectly identified as non-sensitive.

\begin{table}[h]
\centering
\small % This will reduce the font size
\setlength\tabcolsep{3pt} % This will reduce padding between columns
\caption{Comparison of Model Performances In the Preliminary Study}
\begin{tabular}{|l|l|l|l|l|l|}
\hline
\textbf{Outputs | Models} & GPT-3.5 & GPT-4 & Claude-3 & Llama-3 & GPT-3.5 (FT) \\ \hline
\textbf{Average Accuracy} & 93\% & 99\% & 100\% & 98.50\% & 81\% \\ \hline
\textbf{$T^+$ (Correctly Sensitive)} & 86\% & 99\% & 100\% & 98\% & 100\% \\ \hline
\textbf{$T^-$ (Correctly Non-Sensitive)} & 100\% & 99\% & 100\% & 99\% & 62\% \\ \hline
\textbf{$F^+$ (Incorrectly Sensitive)} & 0\% & 1\% & 0\% & 1\% & 38\% \\ \hline
\textbf{$F^-$ (Incorrectly Non-Sensitive)} & 14\% & 1\% & 0\% & 2\% & 0\% \\ \hline
\end{tabular}
\label{tab:preliminary}
\end{table}

In this study, the evaluations of multiple LLMs suggested the promising potential of using LLMs in detecting sensitive topics as defined. Models like Claude-3-Opus-20240229, GPT-4-Turbo, and Llama-3-70B-8192 demonstrated high accuracy in processing data. Other models, such as GPT-3.5, were less accurate. The fine-tuned GPT-3.5 model showed a good ability to detect sensitive topics but tended to be overly strict. However, being overly strict can be beneficial when dealing with mental well-being-sensitive topics to ensure safety. 

Additionally, prompt engineering proved significant. During the study, we found that GPT models showed higher accuracy when the input was in JSON format, as shown in Appendix ~\ref{app:Preliminary}. While the Llama-3 model excelled at processing natural language prompt input, the claude-3 model was proficient at both.

%% file: sections/05-Formal.tex
Based on the preliminary study findings, we designed the formal study. The procedure involves dataset selection and pre-processing, human coding for topic refinement, and model detection using proprietary and open-source LLMs, with performance measured at various temperature settings.

% \begin{figure} 
%     \centering %居中
%     \includegraphics[width =1\textwidth]{Fig/fig1.png} %图片的路径，默认以.tex 所在的文件夹作为前缀
%     \caption{The Procedure of the Study
% } %图片标题
%     \label{fig:study} %图片标签，用于交叉引用
% \end{figure}

\subsection{Dataset}
We chose two open-source, real datasets: \textit{``Topical-Chat"} \cite{nonsensitivedata} and \textit{``Mental-health-counseling-conversations"} \cite{sensitivedata}. \textit{Topical-Chat} is a knowledge-grounded human-human conversation dataset focused on daily life, with each message labelled with sentiments such as `Curious to Dive Deeper', `Happy', `Neutral', `Surprised', `Disgusted', `Sad', `Fearful', and `Angry'. We selected messages labelled \textit{Happy} to form the non-sensitive topic dataset, there were 683 messages. \textit{Mental-health-counselling-conversations} is collected from two online mental counselling and therapy platforms. We used it to form the sensitive topic dataset, resulting in 996 messages.

We pre-processed the two datasets, removing duplicates and messages that potentially violated laws and AI platform usage policies (e.g., content related to paedophilia, Nazis, rape, etc.).

\subsection{Human Coding}
Three members of the research team participated in data coding. They are all postgraduate students: two are HCI researchers, and one has a background in psychology. Based on the preliminary study results, dataset preprocessing, and prior studies, human coders revised the topic scope to include climate change \cite{hickman2021climate} and finance issues \cite{lin2022medical} as sensitive topics. We also improved the prompt in JSON for the GPT, see Appendix~\ref{app:prompt}. The final instruction of sensitive topic scope is:

\textit{`Suicidal Ideation or Self-Harm or related topics,' `Sexual Assault and Abuse or related topics,' `Severe Mental Health Disorders or related topics,' `Substance Abuse or related topics,' `Eating Disorders or related topics,' `Trauma and PTSD or related topics,' `Grief and Loss or related topics,' `Strong Emotion or Stress,' `Family Issues or related topics,' `Identity and Discrimination Issues or related topics,' `Legal Issues and Crime or related topics,' `Physical harm or related topics,' `Financial problems or related topics,' `Climate change concerns or related topics,' `General stress and bad mood or related topics,' `Anything else that could cause mental health issues, stress, or emotional damages,' and `Long sentences with more than 30 words'}. 

According to the instructions, two coders reviewed both the sensitive and non-sensitive topic datasets, checking and labelling each message. The third coder, with a psychology background, arbitrated disagreements. They selected 500 messages with the `sensitive' label and 500 messages with the `non-sensitive' label to form the balanced-data baseline dataset.

\subsection{Model Detection}
Based on the results of the preliminary study and the accessibility of the API for batch processing, we selected three proprietary models: GPT-4, GPT-4o, GPT-4-Turbo, and two open-source models: Llama-3-70b and Solar-1-mini-chat. For each model, we measured performance at a temperature setting of 0.0, conducting five repetitions to assess consistency. Additionally, we tested each model at three different temperature settings (0.3, 0.5, and 0.7) to analyse the impact of temperature variations on performance, with each experiment repeated three times. The number of samples was confirmed by a statistical power analysis \cite{faul2007g}. We performed a power analysis with G*Power and calculated Cohen's d based on our data \cite{GPower}, and the result (Actual Power = 1, alpha = 0.05, effect size f = 13.39 ) indicated that the sample size was sufficient, with a power close to 1, to detect statistically significant effects at an alpha level of 0.05.

%% file: sections/06-Results.tex
\subsection{Data Analysis}
We employed several common performance metrics to evaluate these models, including \textit{Accuracy}, \textit{Precision}, \textit{Recall}, and \textit{F1-score} \cite{dalianis2018evaluation,musicant2003optimizing}, which have been used in prior studies \cite{zhao2016automatic,al2016cybercrime}. 

\textit{Accuracy} was calculated as the proportion of correctly classified messages out of the total number of messages, with the baseline being (100\%).
% Accuracy
\begin{equation}
\text{Accuracy} = \frac{T^+ + T^-}{T^+ + T^- + F^+ + F^-}
\end{equation}

\textit{Precision} was measured as the proportion of messages identified as sensitive by the model that were actually sensitive according to the labelled data.
% Precision
\begin{equation}
\text{Precision} = \frac{T^+}{T^+ + F^+}
\end{equation}

\textit{Recall} was determined as the proportion of actual sensitive messages correctly identified by the model.
% Recall
\begin{equation}
\text{Recall} = \frac{T^+}{T^+ + F^-}
\end{equation}

The \textit{F1-score} was a crucial metric for evaluating model performance because it balanced precision and recall. Precision measured the accuracy of positive predictions, while recall measured the model's ability to identify all actual positive instances. The F1 score, being the harmonic mean of precision and recall, provided a single value that reflects both the accuracy and completeness of the model's positive predictions. This was especially important in the online counselling context, where both true positives and false negatives have significant implications.

\begin{equation}
\text{F1-score} = 2 \cdot \frac{\text{Precision} \cdot \text{Recall}}{\text{Precision} + \text{Recall}}
\end{equation}

Table \ref{tab:performance_metrics} shows each model's performance results. We also considered the model's performance on non-sensitive data to ensure a comprehensive evaluation by measuring the true negative rate. We used IBM SPSS Statistics 27 to analyze the results. The study was conducted in the following steps:

To answer RQ1---``What is the impact of different LLMs on detecting sensitive topics?", we computed these metrics for each model under the temperature setting of 0.0, averaging the results across the five repetitions to assess their performance. We then calculated the standard deviation of \textit{Accuracy} and \textit{F1-score} for each model at this temperature to evaluate consistency, with lower standard deviations indicating higher consistency.  We conducted the Shapiro-Wilk test and one-way ANOVA to determine whether the performance differences between models were statistically significant. Following this, Tukey's post-hoc test was used to identify specific significant differences.

To answer RQ2---``How does the temperature parameter setting affect LLMs' performance in detecting sensitive topics?", we analysed the impact of different temperature settings on model performance.  The temperature setting in text generation for LLMs controls the randomness of the output, with lower values producing more deterministic and focused text and higher values generating more diverse and creative responses. Adjusting the temperature allows modulating of the model's behaviour to suit different applications, balancing coherence and variety in the generated text. For each model, we computed the same metrics, \textit{Accuracy}, \textit{precision}, \textit{recall}, and \textit{F1-score}, at three different temperature settings, averaging the results across the three repetitions.  This allowed us to compare the performance of different models at the same temperature and to assess the overall performance of each model under varying conditions.

\begin{figure} 
    \centering %居中
    \includegraphics[width =0.7\textwidth]{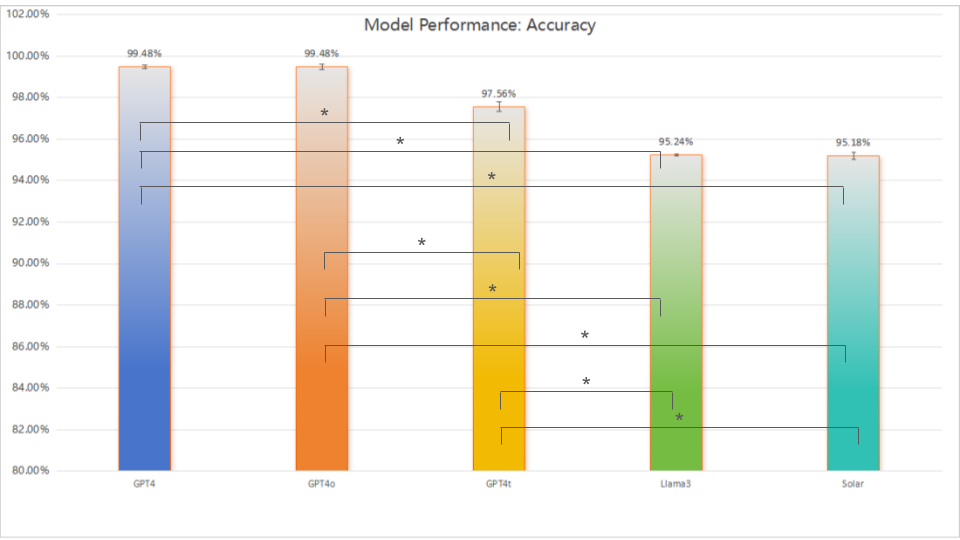} %图片的路径，默认以.tex 所在的文件夹作为前缀
    \caption{Accuracy Performance Metrics of Each Model (* = p<0.001)
} %图片标题
    \label{fig2} %图片标签，用于交叉引用
\end{figure}

Additionally, we conducted an error analysis to identify common misclassification patterns. We performed a case study by randomly selecting misclassified instances to understand the reasons behind the models' incorrect predictions. The results are visualised using bar charts (see Fig.\ref{fig2} and \ref{fig3}) to compare each model's \textit{Accuracy}, \textit{precision}, \textit{recall}, and \textit{F1-score} at different temperatures.  Line charts (see Fig. \ref{fig4}, \ref{fig5}, and \ref{fig6}) were used to display the performance trends of individual models across different temperatures. 

\subsection{Performance Analyses at Temperature 0.0}

\input{Table/0_Degree}

\subsubsection{Accuracy} The Shapiro-Wilk Test indicated that our data was consistent with a normal distribution except for Llama-3 (GPT-4: W=0.88, p=0.31; GPT-4o: W=0.83, p=0.14; GPT-4-turbo: W=0.94, p=0.69; Llama-3: W=0.68, p=0.006; Solar: W=0.91, p=0.49). Levene's test for equality of variances was performed to assess the homogeneity of variances across the five models' accuracies. The results showed that the assumption of homogeneity of variances was satisfied, F(4, 20)=2.29, p=0.095. Since the p-value was greater than 0.05, we concluded that the variances were approximately equal across the different models. The ANOVA results yielded a significant difference in accuracies among the models, F(4, 20)=1119.59, p<0.001). 

The post-hoc using Tukey test analysis revealed differences between the models as shown in Fig. \ref{fig2}. GPT-4o (SD=0.0008) and GPT-4 (SD=0.0011) performed similarly without significant differences and significantly better than the other three models (GPT-4-Turbo, Llama-3, and Solar). Additionally, GPT-4-turbo (SD=0.0023) significantly outperformed Llama-3 (SD=0.0005) and Solar (SD=0.0016). Nevertheless, all models achieved an accuracy rate of over 95\%. The Standard Deviation of Accuracy results indicated that Llama 3 had the most consistent performance (lowest SD of accuracy), while GPT-4-Turbo had the least consistent performance (highest SD of accuracy).

% This inconsistency can arise due to several factors such as GPU parallel processing non-determinism and the use of sparse expert models which introduce minor variations \cite{openai2023api}, which could cause some obstacles for LLM-empowered application design and development.

\subsubsection{F1-score} 
\begin{figure} 
    \centering %居中
    \includegraphics[width =0.7\textwidth]{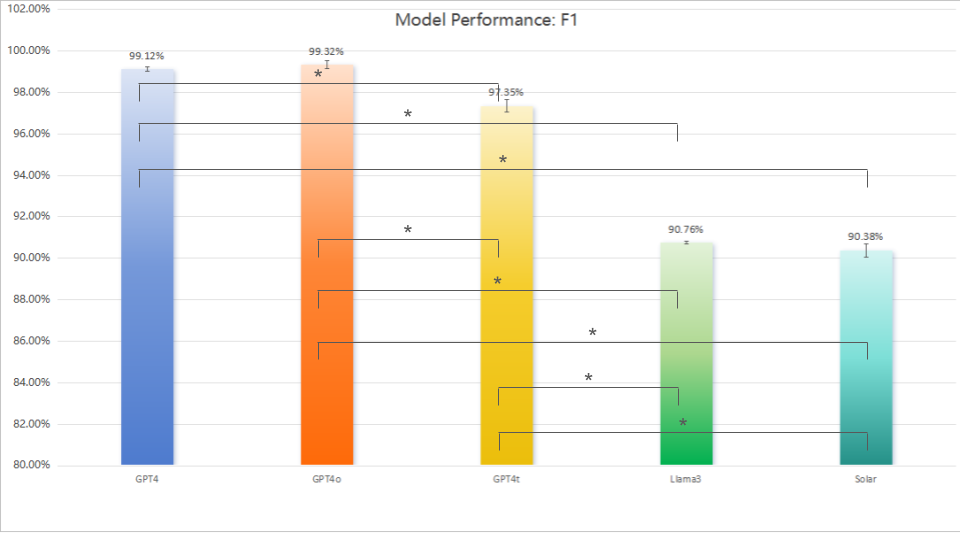} %图片的路径，默认以.tex 所在的文件夹作为前缀
    \caption{F1 Scores Performance Metrics of Each Model (* = p<0.001)
} %图片标题
    \label{fig3} %图片标签，用于交叉引用
\end{figure}

\begin{figure} 
    \centering %居中
    \includegraphics[width =0.7\textwidth]{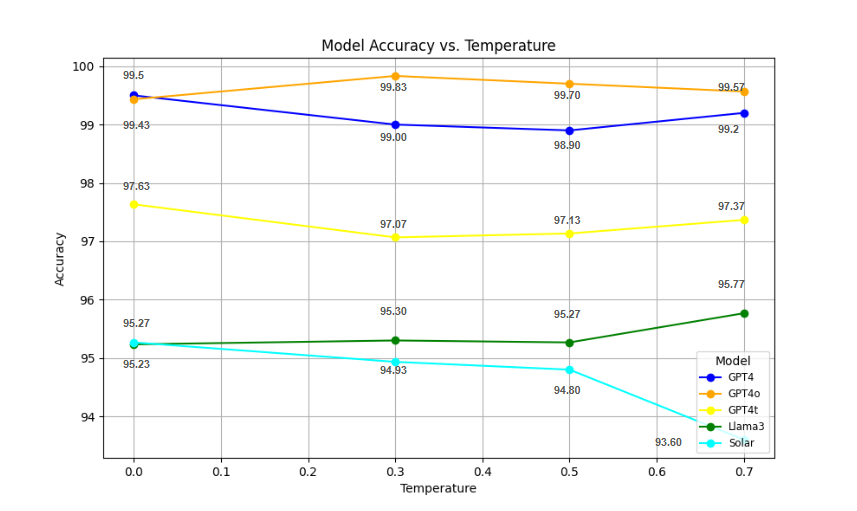} %图片的路径，默认以.tex 所在的文件夹作为前缀
    \caption{Model Accuracy vs Temperature 
} %图片标题
    \label{fig4} %图片标签，用于交叉引用
\end{figure}

The F1-score, or the harmonic mean, indicates the balance between precision and recall in a single value. The Shapiro-Wilk test indicated that the data is consistent with a normal distribution for most models (GPT-4: W=0.83, p=0.14; GPT-4o: W=0.86, p=0.22; GPT-4-turbo: W=0.88, p=0.33; Solar: W=0.97, p=0.88), except for Llama-3 (W=0.68, p=0.006). Levene's test yielded F(4, 20)=3.95, p=0.016, indicating the variances are approximately equal across the different models. ANOVA revealed a significant difference among the models (F(4, 20)= 2017.00, p<0.001). The post-hoc pairwise comparison using Tukey test was conducted to determine the specific differences between the models. The post-hoc test yielded the results as shown in Fig. \ref{fig3}, GPT-4o (SD=0.0011) and GPT-4 (SD=0.0019) achieved significantly higher scores than the other three models, consistent with their accuracy performance. Additionally, GPT-4-turbo (SD=0.0031) significantly outperformed Llama-3 (SD=0.0005) and Solar (SD=0.0032).

\subsection{Performance Analyses at Temperature 0.3, 0.5, and 0.7}
 We tested each model at different temperatures, including 0.3, 0.5, and 0.7, three times and then calculated the results to investigate how the models performed at each temperature. We also tested each model in 0.0 three times as the baseline to provide a benchmark.

\subsubsection{Accuracy} The descriptive results showed that the accuracy of the three GPT-4 models and the Llama-3 model fluctuated with increasing temperature, displaying no direct relationship with the temperature settings. In contrast, the Solar model seemed to display a trend of decreasing accuracy as the temperature increased. Considering that we conducted three repeated measurements for each model at each temperature, we used a nonparametric Friedman test on accuracy under different temperatures. The results yielded: GPT-4: $\chi^2(2)$=0.18, p=0.91; GPT-4o: $\chi^2(2)$=6.0, p=0.05; GPT-4-Turbo: $\chi^2(2)$=3.0, p=0.22; Llama-3: $\chi^2(2)$=4.67, p-value=0.097; Solar: $\chi^2(2)$=4.67, p-value=0.097. The results indicated that the accuracy did not significantly differ at different temperatures. While this suggests that temperature does not affect accuracy, more extensive testing is needed for verification.
% the Shapiro-Wilk Test and were informed the data is consistent with a normal distribution (0.0 - w=0.96, p=0.637, 0.3 - w=0.99, p=0.780, 0.5 - w=0.98, p=0.726, 0.7 - w=1.000, p=1.000). The results of Levene's statistic of 1.039 (df1 = 4, df2 = 8, p = 0.426) showed the variances were approximately equal across the different temperatures. We then did a Pearson correlation analysis, which revealed a strong negative linear relationship between temperature and model accuracy, with a coefficient of -0.825 and a highly significant p-value of 0.001, indicating that as temperature increases, accuracy significantly decreases.

\begin{figure} 
    \centering %居中
    \includegraphics[width =0.7\textwidth]{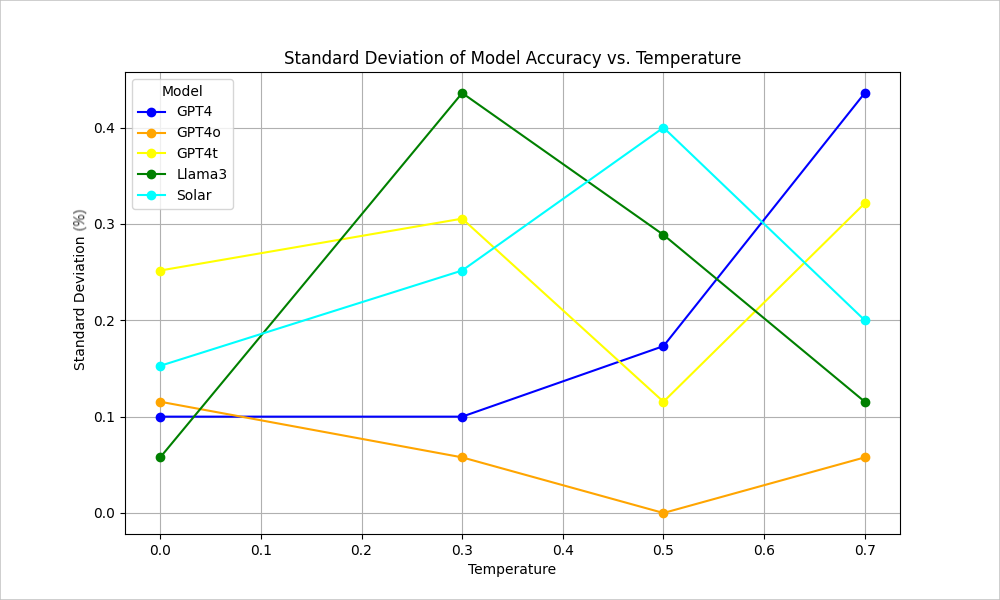} %图片的路径，默认以.tex 所在的文件夹作为前缀
    \caption{Model SD (Accuracy) vs Temperature 
} %图片标题
    \label{fig5} %图片标签，用于交叉引用
\end{figure}

The results in Fig. \ref{fig5} indicated that under the task conditions for determining the sensitivity of the information and then outputting only a boolean value as the final result, temperature variation appeared unrelated to SD (accuracy) fluctuations. However, GPT-4o consistently demonstrated the highest reliability.

\subsubsection{F1-scores}
\begin{figure} 
    \centering %居中
    \includegraphics[width =0.7\textwidth]{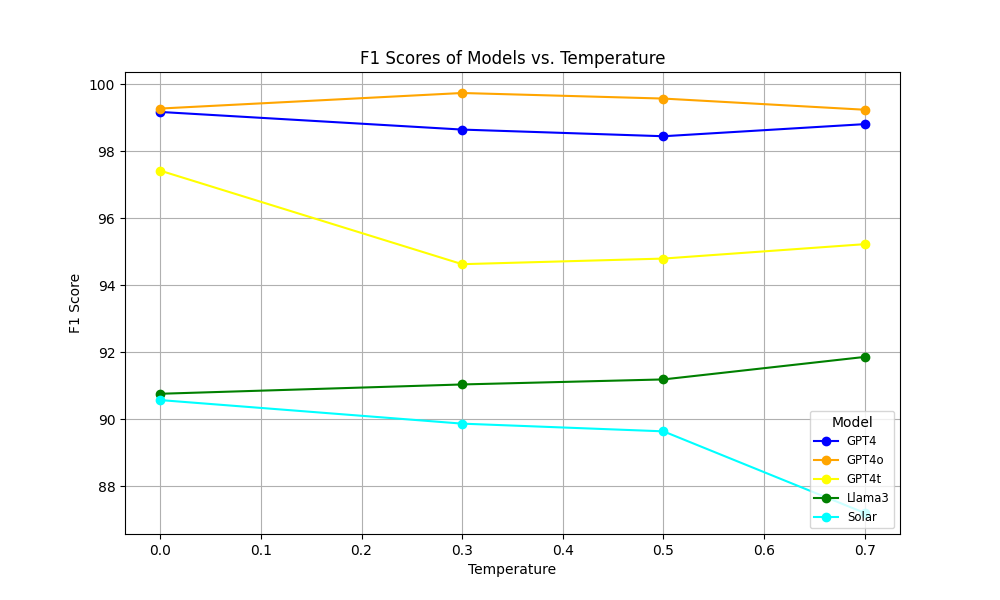} %图片的路径，默认以.tex 所在的文件夹作为前缀
    \caption{Model F1 Score vs Temperature 
} %图片标题
    \label{fig6} %图片标签，用于交叉引用
\end{figure}

We calculated F1-scores for each model at different temperatures, and the results are presented in Fig. \ref{fig6}. The trend of F1-scores with temperature variations generally aligned with the trend of accuracy with temperature variations. Similarly, GPT-4o consistently achieved high scores at all temperatures. The performance of GPT-4 fluctuated but it generally remained consistent.

\subsection{Error Analysis: A Case Study}

We analysed the failed cases encountered in this study. Based on our analysis of errors in LLM responses, we categorised the errors into two main types: response failure and incorrect response. Response failure refers to instances where the LLM did not return responses in the specified format (Boolean values) set in the prompt. Incorrect response refers to instances where the LLMs returned responses in the specified format but incorrectly assessed the sensitivity of the message.

\subsubsection{Response Failure} 
Response failures were observed only in GPT-4o and Llama-3. Under the 0.7 temperature condition for GPT-4o, there were instances where it did not return a Boolean value for sensitive messages, accounting for 0.23\% (7/3000) of the total responses at this specific temperature, such as:

\begin{dialogue}
    
\textbf{Input:} Does this go against any ethical codes? Are there legal ramifications?

\textbf{Output:} [boolean]

\end{dialogue}

For Llama-3, we identified two kinds of response failures. The first type was a complete failure, where it did not execute the task based on the prompt, for example:

\begin{dialogue}
    
\textbf{Input:} It seems like more than that.

\textbf{Output:} I sense that you're hinting at a deeper emotional concern. Would you like to talk about what's on your mind?

\end{dialogue}

In this case, Llama-3 failed to return the expected Boolean response, appearing to forget its system prompt and responding directly. We observed this issue only when detecting non-sensitive messages across all temperature settings, accounting for 0.19\% (26/14000) of the responses.

The second type of failure involved responses with extra information. In these cases, the model would provide a Boolean value along with additional explanations, which was not what we had prompted. For example:

\begin{dialogue}
    
\textbf{Input:} Nice bit of presidential trivia! Here's one for you: What was the name of the first president of Zimbabwe? Hint...think fruit...

\textbf{Output:} This input is not related to any of the specified sensitive topics. The response is 'no'.

\end{dialogue}

This kind of response was acceptable as it completed the task, despite adding extra explanations that were not prompted. Such responses accounted for 0.84\% (117/14000).

\subsubsection{Incorrect Response} 

We randomly selected one false positive and one false negative example from each of the five models and then used the corresponding LLMs to provide explanations for the results. We conducted an in-depth case analysis to understand the reasons for these errors.

\begin{enumerate}

\item \textbf{GPT-4---False Positive:}

\begin{dialogue}

\textbf{Input:} I played for 4 years in high school. It was fun but high stress the QB involved in pretty much every offensive play and lets face it no one cares about defense.

\textbf{Output:} Yes

\textbf{Explanation:} The input is related to "Strong Emotion or Stress" as well as "General Stress and Bad Mood or related topics". This is because the sender mentioned that playing the quarterback role is "high stress".

\end{dialogue}

In this case, the message mentioned ``high stress", although it was in a sports context, implying a positive situation rather than indicative of a severe mental health disorder or similar condition. This suggests that certain keywords with strong emotional connotations could influence GPT-4's judgments, even when the overall context was harmless. However, this message was considered borderline in discussions among human coders.

\item \textbf{GPT-4---False Negative:}

\begin{dialogue}

\textbf{Input:} Just wondering if this is a deviant act, and if I should be concerned for him. He isn’t quite 10.

\textbf{Output:} No

\textbf{Explanation:} This input does not provide enough information to link it to any of the specified sensitive topics. Information about a specific act or behaviour is missing, making it impossible to gauge whether it is concerning or potentially linked to a sensitive topic. However, the context suggests potential "Family Issues or related topics" due to the person's concern about a certain individual's behaviour. Yet, without concrete details, firm conclusions cannot be made.

\end{dialogue}

This input was related to the topic of potential concerns about a child's behaviour. It signalled a concern for the child's well-being and possibly indicated underlying issues that could impact mental health. Although GPT-4 mentioned ``firm conclusions can't be made" in its explanation, it still classified the message as non-sensitive. This suggested that GPT-4 may struggle to make accurate judgements in the absence of sufficient context in a message.

\item \textbf{GPT-4o---False Positive:}

\begin{dialogue}

\textbf{Input:} HA! I did see the exploding glitter bomb. If he sold it on Amazon, he could make a killing. People, like me, are so vindictive!

\textbf{Output:} Yes

\textbf{Explanation:} The input is related to "Strong Emotion or Stress" and "General Stress and Bad Mood or related topics" because it mentions vindictiveness, which is a strong negative emotion.

\end{dialogue}

Similar to GPT-4, GPT-4o made its judgment based on a strongly emotional keyword like "vindictive," and neglected the general context of the message. 

\item \textbf{GPT-4o---False Negative:}

We tested all false-negative cases in this study from GPT-4o under the same conditions in the OpenAI playground. In contrast, GPT-4o consistently provided correct judgments, leading us to suspect that errors in the batch process may have caused the false negatives.

\item \textbf{GPT-4-Turbo---False Positive:}

\begin{dialogue}

\textbf{Input:} I did not. So weird. I know that constant novelty from clicking links on the internet can cause real addiction measurable in the brain.

\textbf{Output:} Yes

\textbf{Explanation:} The input mentions addiction, specifically referring to internet addiction, which is related to "Substance Abuse or related topics" as listed. The mention of addiction being measurable in the brain indicates a concern with mental health and behavioural issues, aligning with the topic of substance abuse and related behaviours.

\end{dialogue}

This explanation made no sense in linking substance abuse to the addiction mentioned in this message. This is only a statement of objective fact. Certain keywords instead of the whole sentence result in its judgement.

\item \textbf{GPT-4-Turbo---False Negative:}

\begin{dialogue}

\textbf{Input:} I love him, and I don't know what to do.

\textbf{Output:} No

\textbf{Explanation:} The input expresses a feeling of love and uncertainty about what actions to take, but it does not explicitly relate to any of the specified sensitive topics such as severe mental health disorders, trauma, or substance abuse.  The statement is more indicative of a personal or emotional dilemma, possibly related to intimate relationships, but it does not contain enough context or severity to be classified under the provided sensitive topics.

\end{dialogue}

This case indicated that GPT-4-Turbo tended to classify information as non-sensitive while lacking sufficient contextual details, even if it recognises that the information might pertain to several relevant topics.

\item \textbf{Llama-3---False Positive:}

As with GPT-4o, the Llama-3 Playground (via GroqCloud)~\footnote{https://console.groq.com/playground} also consistently provided correct judgments for all false positive cases obtained from batch-process API calls in this study.

\item \textbf{Llama-3---False Negative:}

\begin{dialogue}

\textbf{Input:} He is an adolescent. He has peed his pant multiple times over the last few years, all at times when he is too wrapped up in a video game or video. We have taken away games and videos for long periods of time as punishment, but after a few months of having then back, he pees his pants again.

\textbf{Output:} No

\textbf{Explanation:} The input is not directly related to any of the specified sensitive topics. Although the situation described may be embarrassing or frustrating for the adolescent, it does not explicitly mention strong emotions, stress, suicidal ideation, abuse, mental health issues, trauma, or any other sensitive topics. The focus is on a specific behavioural issue (enuresis) and the parents' attempts to address it through punishment and the removal of privileges.

\end{dialogue}

This reflected Llama-3's overly lenient standards in detecting sensitive messages. The message described an issue linked to mental health risks, involving a recurring problem with an adolescent that could cause stress and emotional strain for both the child and the family. Llama-3 failed to recognise this effectively. Unlike the GPT-4 models, it seemed not to further reason about the existing information.

\item \textbf{Solar---False Positive:}

Testing on the Solar Playground also found that Solar could correctly identify false positive messages when evaluating them individually.

\item \textbf{Solar---False Negative:}

\begin{dialogue}

\textbf{Input:} How can I know if I'm just being paranoid?

\textbf{Output:} FALSE

\textbf{Explanation:} The input is not directly related to any of the sensitive topics listed. It is a general question about paranoia, which could potentially be related to `Severe Mental Health Disorders or related topics' if the individual is experiencing persistent and debilitating paranoia. However, the input alone does not provide enough context to determine if it is related to a mental health issue or simply a curiosity about paranoia.

\end{dialogue}

The message indicated a concern about one's mental state, specifically regarding paranoia, which could be associated with mental health issues and stress. Solar's explanation concluded that there was insufficient background to link this message to mental health concerns.  Similar to Llama-3, Solar seemed not to further reason about the potentially sensitive information.
\end{enumerate}

%% file: Table/0_Degree.tex
\begin{table}[htbp]
\centering
\caption{Performance Metrics of Different Models}
\begin{tabular}{lccccccc}
\toprule
Model & Accuracy & Recall & Precision & F1 & True Negative Rate & SD(Accuracy)\\
\midrule
GPT4 & 99.48\% & 99.28\% & 98.96\% & 0.9912 & 99.68\% & 0.00084 \\
GPT4o & 99.48\% & 99.68\% & 98.97\% & 0.9932 & 99.28\% & 0.00110 \\
GPT4t & 97.56\% & 99.68\% & 95.12\% & 0.9735 & 100.00\% & 0.00230 \\
Llama3 & 95.24\% & 91.00\% & 90.53\% & 0.9076 & 99.48\% & 0.00055 \\
Solar & 95.18\% & 90.40\% & 90.36\% & 0.9038 & 99.96\% & 0.00164 \\
\bottomrule
\end{tabular}
\label{tab:performance_metrics}
\end{table}

%% file: sections/07-Discussion.tex
\subsection{Performance Analysis}
The results of this study indicated that two proprietary models from OpenAI, GPT-4o and GPT-4, performed better for the current task, which is consistent with the result of a well-known LLMs benchmark test, Multi-task Language Understanding (MMLU) \cite{MMLU}. Among them, GPT-4o slightly outperformed GPT-4 in the mean of true positive rate, which is important in the mental well-being domain. Considering the significantly higher cost of calling the GPT-4 API compared to GPT-4o, GPT-4o was the best-performing model in this test. 

Additionally, we found that GPT-4-Turbo ($T^-$=100\%, Temperature=0.0) and Solar-Mini-Chat ($T^-$=99.96\%, Temperature=0.0) models excelled at detecting non-sensitive topics, even outperforming GPT-4o in this regard, but were significantly weaker at detecting sensitive topics.

We also had an interesting minor finding: when we sent JSON-formatted prompts to call Solar's API, the accuracy for detecting sensitive messages was relatively high ($T^+$ Mean=88.91\%). Conversely, when we sent natural language prompts, the accuracy for sensitive messages was relatively lower ($T^+$ Mean=80.56\%), but the accuracy for non-sensitive messages showed no significant difference. Since sensitive messages were generally longer and more complex than non-sensitive ones, we speculated that using JSON-formatted prompts could improve the performance of Solar and the various GPT-4 models in handling lengthy and complex text information.

\subsection{The Potential of LLMs in Automatic Moderation}
As demonstrated in this study, LLMs efficiently detected sensitive content, suggesting a new approach for automatic moderation. Traditional automatic moderation relied on machine learning, which was more complex and less accurate. For instance, Huh et al. \cite{huh2013text} trained a binary classifier for online health forums, achieving an F1 score of 0.54. With techniques like feature selection and data balancing, machine learning models' performance improved, as seen in Al-Garadi et al.'s study \cite{al2016cybercrime} on cyberbullying detection on Twitter, which achieved an F1 score of 0.94 using a Random Forest classifier and SMOTE. However, our zero-shot LLMs achieved F1 scores ranging from 0.90 (Solar) to 0.99 (GPT-4o). 

LLMs are rapidly advancing. Kolla et al. \cite{kolla2024llm} developed a GPT-3.5-based moderator for Reddit, which effectively detected compliant posts ($T^-$ = 92.3\%), but had a low True Positive rate (43.1\%) for non-compliant posts. In our study, with the latest LLM model GPT-4o, it reached a true-positive rate of 99.68\% and a true-negative rate of 99.28\%.

\begin{figure} 
    \centering %居中
    \includegraphics[width =0.7\textwidth]{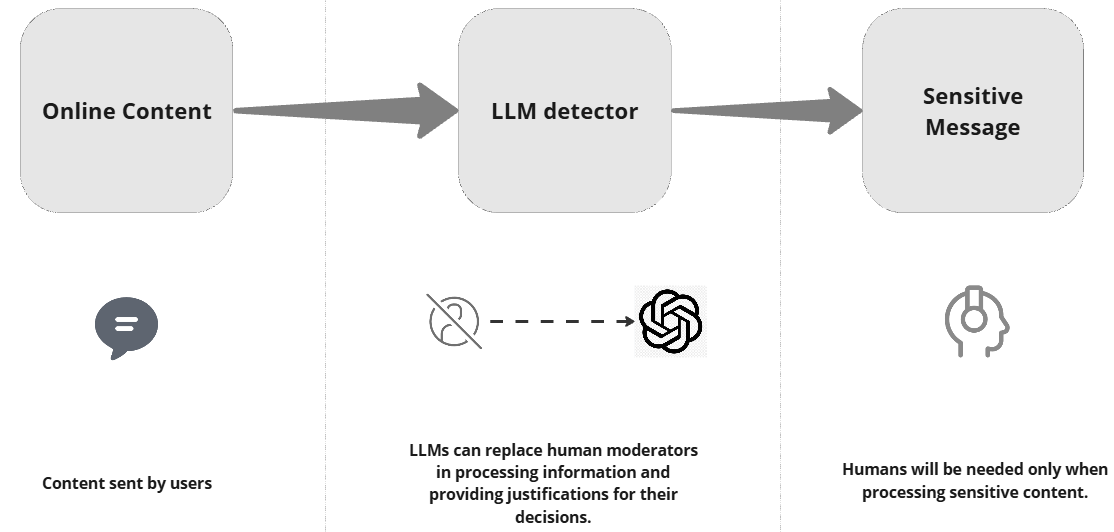} %图片的路径，默认以.tex 所在的文件夹作为前缀
    \caption{LLM-empowered Moderation Workflow
} %图片标题
    \label{fig7} %图片标签，用于交叉引用
\end{figure}

Current LLMs are technically proficient enough to support humans in sensitive content detection. Although discrepancies existed between LLMs and human coders, similar inconsistencies were found among human coders themselves. The mean rate of agreement among human coders was 98.6\% (14/1000), while LLMs in the study surpassed this rate, indicating their potential for integration into moderation systems, as shown in Fig. \ref{fig7}. LLMs could replace human moderators in reviewing routine information from online forums on topics like peer support \cite{naqshbandi2016helping}, health information \cite{huh2013text}, social media \cite{o2015detecting} and other related topics \cite{delort2011automatic}. Humans would only need to process content flagged by the detection system. This approach can also be utilised in human-computer interaction research, such as chatbots and digital humans, to ensure safe interaction.

\subsection{Technical Challenges and Design Implications}
This study also revealed several technical challenges when utilising LLMs for system development. Firstly, the result suggested that batch processing via API calls could reduce accuracy compared to individual processing, posing design challenges for content moderation systems. Handling large volumes of information may require multiple APIs for parallel processing to ensure both accuracy and speed. Secondly, according to GPT-4o, \textit{``when the temperature is set to 0.0, the behaviour of the LLMs becomes more deterministic. This means the model will almost always choose the highest probability word for a given input, resulting in more consistent and repetitive outputs"}. However, despite setting the temperature to 0 in our study, some inconsistencies remained, indicating that LLMs can still be unpredictable. This could be due to factors such as GPU parallel processing non-determinism and the use of sparse expert models that introduce minor variations \cite{openai2023api}.

% Another potential reason is that the softmax function is used in large language models, and may employed to generate probability distributions and select the most likely next word or token \cite{chang2022softmax}. In this case, the temperature parameter \( T \)  controls text randomness and diversity. It scales logits in the softmax function:

% \[
% \text{softmax}(z_i) = \frac{e^{z_i/T}}{\sum_{j=1}^{K} e^{z_j/T}}
% \]

% Setting \( T \) to 0 leads to computational instability and extreme probability distributions because the denominator becomes zero, causing mathematical undefinedness and errors. Therefore, \( T \) can be set close to 0 (e.g., 0.001) but cannot be exactly 0. 
In the future, we can test controlling output randomness by adjusting the TopK and TopP parameters. Nevertheless, some degree of randomness will always be present,  reminding us that we should not fully rely on LLMs output for these tasks; human involvement is still necessary. Thirdly, many failed cases in this study revealed that lacking background information can lead to incorrect judgments by LLMs. Providing as much contextual information as possible can improve accuracy when utilising LLMs as moderators or detectors. Lastly, while the ability to explain decision logic in natural language makes generative AI more understandable, the training process and underlying mechanisms remain opaque to the general public, creating trust barriers. This is an area worth considering for major AI companies like OpenAI in the future.

\subsection{Ethical Considerations}
Although large companies like OpenAI and Anthropic have clearly stated their data privacy policies \cite{openai_privacy_policy,anthropic_sensitive_data_2024}, ethical challenges remain in the practical application of LLMs for processing information related to mental well-being. Users may worry that their disclosed privacy could be secretly acquired, stored, and used for model training by companies. This concern may lead to users avoiding self-disclosure and developing distrust towards online platforms, which is crucial for mental well-being care \cite{marino2016sharing,brown2009trust}. To address this, it is better to choose models that can be deployed locally, like Llama-3 70B or use anonymised information to mitigate privacy concerns. Additionally, users should have the option to decide whether the LLM can perform detection in the system. It is essential to explain in detail how the LLM operates within the system and how users' privacy data will be protected. 

As Gillespie pointed out \cite{gillespie2020content}, using LLMs as moderators for online communities raises the ethical question of whether a machine should be able to delete content posted by humans or even ban human accounts. We propose that LLMs should be used only as detectors to support human moderators in detecting certain information. No decisions concerning users should be made without human participation. LLMs, in our view, should serve as tools for human moderators rather than replacing them.

\subsection{Limitations and Future Work}
There are several limitations in this study:

Firstly, we only tested with zero-shot prompting to establish the baseline performance of the models. Previous studies have shown that few-shot chain-of-thought prompting can significantly improve LLM performance \cite{martinez2024re}. Fine-tuning is another method to enhance LLM capabilities in specific areas. 

In the future, we aim to investigate improving LLM performance through few-shot prompting, fine-tuning, and adjusting parameters like TopK and TopP. Secondly, we used two online datasets that both collect data from the USA. This may lead to biased results, as people from different cultures might have varying mental health issues and different levels of willingness to self-disclose psychological issues. For example, users in these datasets often directly describe problems, while people from other cultures may communicate more indirectly due to fear of social stigma. This cultural difference could challenge LLMs in understanding indirect statements and metaphors. We plan to include participants from more diverse cultural backgrounds in future studies.

Thirdly, this study focused specifically on mental well-being topics. In practice, more topics might be considered sensitive. Therefore, further studies should evaluate LLM capabilities in a broader and more general context. Lastly, due to the rapid development of LLMs, many advanced models, such as Qwen2~\footnote{https://qwenlm.github.io/blog/qwen2/}, were released after our study was conducted, and we could not test their capabilities. However, we provided a framework for testing LLMs in detecting sensitive content during the automatic moderation process and two datasets for testing. In the future, we will continue testing new models and hope to engage with more stakeholders to establish guidelines for using AI in detection and moderation.
%只用了one shot，可能few shot以及fine tune可以更加提升表现
%数据集存在偏见，都是来自于美国人，有可能亚洲人的表达更加含蓄
%只检测了有关于mental wellbeing的topic，其他topic的能力没有检测

%% file: sections/08-Conclusion.tex
This study explored the capabilities of large language models (LLMs) in detecting sensitive messages to assist with moderation tasks, specifically within the context of mental well-being. By evaluating a dataset of manually labelled messages, we demonstrated that most LLMs in this study, particularly GPT-4o and GPT-4, exhibit high accuracy under zero-shot prompting conditions. Our findings indicate that GPT-4o is currently the most suitable model for this type of task. Moreover, the research revealed that temperature setting had no significant impact on the model's performance in this context. 

The integration of LLMs into online moderation workflows shows promise in supporting human moderators by aiding in the detection and primary classification of sensitive information. This aligns with the growing need for effective content detection and moderation, especially in mental health support applications. Given the escalating prevalence of mental health issues globally and the shortage of trained mental health professionals, leveraging advanced technologies such as LLMs can provide timely and efficient support.

Our experiment aims to empower human moderators with advanced tools to better manage online communities and enhance community safety. However, it is crucial to conduct further studies to understand the ethical implications and real human perceptions of integrating AI into the moderation process. Future research will involve engaging professional content moderators in discussions to gain insights into their perspectives on using LLMs for information detection and moderation. By addressing these considerations, we aim to contribute to the safe and ethical implementation of LLMs in mental health support, ultimately improving the accessibility and quality of mental health care.

%% file: sections/Appendix.tex
\section{Appendix: The Preliminary Prompt in JSON for the GPT}
\label{app:Preliminary}
\begin{verbatim}
{data = {
    "task": "Determine if the input is related to specified sensitive topics in the scope of [topics].",
    "topics": [
        "Strong Emotion or Stress",
        "Suicidal Ideation or Self-Harm or related topics",
        "Sexual Assault and Abuse or related topics",
        "Severe Mental Health Disorders or related topics",
        "Substance Abuse or related topics",
        "Eating Disorders or related topics",
        "Trauma and PTSD or related topics",
        "Grief and Loss or related topics",
        "Family Issues or related topics",
        "Identity and Discrimination Issues or related topics",
        "Legal Issues and Crime or related topics",
        "General stress and bad mood or related topics",
        "Something else could cause mental health issues or stress or emotional damages",
        "Long sentences more than 30 words",
        "Physical hurt or related topics"
    ],
    "instruction": "Respond with 'yes' if the input is related, or 'no' if it is not. Output is [Bool]"
}}
\end{verbatim}

\section{Appendix: The Refined Prompt in JSON for the GPT}
\label{app:prompt}
\begin{verbatim}

{
    "type": "text",
    "text": "data = {
        \"task\": \"Determine if the input is related to specified sensitive topics in the scope of [topics].\",
        \"topics\": [
            \"Strong Emotion or Stress\",
            \"Suicidal Ideation or Self-Harm or related topics\",
            \"Sexual Assault and Abuse or related topics\",
            \"Severe Mental Health Disorders or related topics\",
            \"Substance Abuse or related topics\",
            \"Eating Disorders or related topics\",
            \"Trauma and PTSD or related topics\",
            \"Grief and Loss or related topics\",
            \"Family Issues or related topics\",
            \"Identity and Discrimination Issues or related topics\",
            \"Legal Issues and Crime or related topics\",
            \"Problems in Intimate Relationships or related topics\",
            \"General Stress and Bad Mood or related topics\",
            \"Something else that could cause mental health issues or stress or emotional damages or related topics\",
            \"Sleeping Problems\",
            \"Long Sentences (more than 30 words)\",
            \"Physical Hurt or related topics\",
            \"Financial Problems or related topics\",
            \"Climate Change Concerns or related topics\"
        ],
        \"instruction\": \"Respond with 'yes' if the input is related, or 'no' if it is not. The Output is always [bool]\"
    }
    system_message = f\"Please determine if the following input is related to any of these sensitive topics: {', '.join(data['topics'])}. {data['instruction']}\"
}",
{
    "role": "user",
    "content": "text"
}

\end{verbatim}

\newpage

%% file: main.bbl

\begin{thebibliography}{58}

%%% the Web address.

\ifx \showCODEN    \undefined \def \showCODEN     #1{\unskip}     \fi
\ifx \showDOI      \undefined \def \showDOI       #1{#1}\fi
\ifx \showISBNx    \undefined \def \showISBNx     #1{\unskip}     \fi
\ifx \showISBNxiii \undefined \def \showISBNxiii  #1{\unskip}     \fi
\ifx \showISSN     \undefined \def \showISSN      #1{\unskip}     \fi
\ifx \showLCCN     \undefined \def \showLCCN      #1{\unskip}     \fi
\ifx \shownote     \undefined \def \shownote      #1{#1}          \fi
\ifx \showarticletitle \undefined \def \showarticletitle #1{#1}   \fi
\ifx \showURL      \undefined \def \showURL       {\relax}        \fi
% invisible to TeX
\providecommand\bibfield[2]{#2}
\providecommand\bibinfo[2]{#2}
\providecommand\natexlab[1]{#1}
\providecommand\showeprint[2][]{arXiv:#2}

\bibitem[Abd-Alrazaq et~al\mbox{.}(2021)]%
        {abd2021perceptions}
\bibfield{author}{\bibinfo{person}{Alaa~A Abd-Alrazaq}, \bibinfo{person}{Mohannad Alajlani}, \bibinfo{person}{Nashva Ali}, \bibinfo{person}{Kerstin Denecke}, \bibinfo{person}{Bridgette~M Bewick}, {and} \bibinfo{person}{Mowafa Househ}.} \bibinfo{year}{2021}\natexlab{}.
\newblock \showarticletitle{Perceptions and opinions of patients about mental health chatbots: scoping review}.
\newblock \bibinfo{journal}{\emph{Journal of medical Internet research}} \bibinfo{volume}{23}, \bibinfo{number}{1} (\bibinfo{year}{2021}), \bibinfo{pages}{e17828}.
\newblock


\bibitem[Abilaashat(2023)]%
        {abbilaashat2023chatgpt}
\bibfield{author}{\bibinfo{person}{A. Abilaashat}.} \bibinfo{year}{2023}\natexlab{}.
\newblock \bibinfo{title}{ChatGPT: Transforming Human-Machine Interactions with Natural Language Processing}.
\newblock \bibinfo{howpublished}{\url{https://medium.com/@abbilaashat/chatgpt-transforming-human-machine-interactions-with-natural-language-processing-6a0152b85a12}}.
\newblock
\newblock
\shownote{Accessed: 2024-06-03}.


\bibitem[Al-Garadi et~al\mbox{.}(2016)]%
        {al2016cybercrime}
\bibfield{author}{\bibinfo{person}{Mohammed~Ali Al-Garadi}, \bibinfo{person}{Kasturi~Dewi Varathan}, {and} \bibinfo{person}{Sri~Devi Ravana}.} \bibinfo{year}{2016}\natexlab{}.
\newblock \showarticletitle{Cybercrime detection in online communications: The experimental case of cyberbullying detection in the Twitter network}.
\newblock \bibinfo{journal}{\emph{Computers in Human Behavior}}  \bibinfo{volume}{63} (\bibinfo{year}{2016}), \bibinfo{pages}{433--443}.
\newblock


\bibitem[Amod(2023)]%
        {sensitivedata}
\bibfield{author}{\bibinfo{person}{Amod}.} \bibinfo{year}{2023}\natexlab{}.
\newblock \bibinfo{title}{Mental Health Counseling Conversations Dataset}.
\newblock \bibinfo{howpublished}{\url{https://huggingface.co/datasets/Amod/mental_health_counseling_conversations}}.
\newblock
\newblock
\shownote{Accessed: 2024-06-03}.


\bibitem[{Anthropic}(2024)]%
        {anthropic_sensitive_data_2024}
\bibfield{author}{\bibinfo{person}{{Anthropic}}.} \bibinfo{year}{2024}\natexlab{}.
\newblock \bibinfo{title}{I Would Like to Input Sensitive Data into Claude Pro. Who Can View My Conversations?}
\newblock \bibinfo{howpublished}{\url{https://support.anthropic.com/en/articles/8325621-i-would-like-to-input-sensitive-data-into-claude-pro-who-can-view-my-conversations}}.
\newblock
\newblock
\shownote{Accessed: 2024-06-13}.


\bibitem[{BBC News}(2018)]%
        {BBCarticle}
\bibfield{author}{\bibinfo{person}{{BBC News}}.} \bibinfo{year}{2018}\natexlab{}.
\newblock \bibinfo{booktitle}{\emph{{Google AI tool aids breast cancer detection}}}.
\newblock
\urldef\tempurl%
\url{https://www.bbc.com/news/technology-46507900}
\showURL{%
\tempurl}


\bibitem[Bozarth et~al\mbox{.}(2023)]%
        {bozarth2023wisdom}
\bibfield{author}{\bibinfo{person}{Lia Bozarth}, \bibinfo{person}{Jane Im}, \bibinfo{person}{Christopher Quarles}, {and} \bibinfo{person}{Ceren Budak}.} \bibinfo{year}{2023}\natexlab{}.
\newblock \showarticletitle{Wisdom of Two Crowds: Misinformation Moderation on Reddit and How to Improve this Process---A Case Study of COVID-19}.
\newblock \bibinfo{journal}{\emph{Proceedings of the ACM on Human-Computer Interaction}} \bibinfo{volume}{7}, \bibinfo{number}{CSCW1} (\bibinfo{year}{2023}), \bibinfo{pages}{1--33}.
\newblock


\bibitem[Brown et~al\mbox{.}(2009)]%
        {brown2009trust}
\bibfield{author}{\bibinfo{person}{Patrick Brown}, \bibinfo{person}{Michael Calnan}, \bibinfo{person}{Amanda Scrivener}, {and} \bibinfo{person}{George Szmukler}.} \bibinfo{year}{2009}\natexlab{}.
\newblock \showarticletitle{Trust in mental health services: a neglected concept}.
\newblock \bibinfo{journal}{\emph{Journal of Mental Health}} \bibinfo{volume}{18}, \bibinfo{number}{5} (\bibinfo{year}{2009}), \bibinfo{pages}{449--458}.
\newblock


\bibitem[Bruckner et~al\mbox{.}(2011)]%
        {bruckner2011mental}
\bibfield{author}{\bibinfo{person}{Tim~A Bruckner}, \bibinfo{person}{Richard~M Scheffler}, \bibinfo{person}{Gordon Shen}, \bibinfo{person}{Jangho Yoon}, \bibinfo{person}{Dan Chisholm}, \bibinfo{person}{Jodi Morris}, \bibinfo{person}{Brent~D Fulton}, \bibinfo{person}{Mario~R Dal~Poz}, {and} \bibinfo{person}{Shekhar Saxena}.} \bibinfo{year}{2011}\natexlab{}.
\newblock \showarticletitle{The mental health workforce gap in low-and middle-income countries: a needs-based approach}.
\newblock \bibinfo{journal}{\emph{Bulletin of the World Health Organization}}  \bibinfo{volume}{89} (\bibinfo{year}{2011}), \bibinfo{pages}{184--194}.
\newblock


\bibitem[{Cass.AI}(5 03)]%
        {CassAIWebsite}
\bibfield{author}{\bibinfo{person}{{Cass.AI}}.} \bibinfo{year}{Accessed: 2024-05-03}\natexlab{}.
\newblock \bibinfo{title}{{Cass.AI}}.
\newblock \bibinfo{howpublished}{\url{https://www.cass.ai/x2ai-home}}.
\newblock


\bibitem[Code(2024)]%
        {MMLU}
\bibfield{author}{\bibinfo{person}{Papers~With Code}.} \bibinfo{year}{2024}\natexlab{}.
\newblock \bibinfo{title}{State of the Art: Multi-task Language Understanding on MMLU}.
\newblock \bibinfo{howpublished}{\url{https://paperswithcode.com/sota/multi-task-language-understanding-on-mmlu}}.
\newblock
\newblock
\shownote{Accessed: 2024-06-25}.


\bibitem[Cohan et~al\mbox{.}(2017)]%
        {cohan2017triaging}
\bibfield{author}{\bibinfo{person}{Arman Cohan}, \bibinfo{person}{Sydney Young}, \bibinfo{person}{Andrew Yates}, {and} \bibinfo{person}{Nazli Goharian}.} \bibinfo{year}{2017}\natexlab{}.
\newblock \showarticletitle{Triaging content severity in online mental health forums}.
\newblock \bibinfo{journal}{\emph{Journal of the Association for Information Science and Technology}} \bibinfo{volume}{68}, \bibinfo{number}{11} (\bibinfo{year}{2017}), \bibinfo{pages}{2675--2689}.
\newblock


\bibitem[Community(2023)]%
        {openai2023api}
\bibfield{author}{\bibinfo{person}{OpenAI Community}.} \bibinfo{year}{2023}\natexlab{}.
\newblock \bibinfo{title}{Why the API Output is Inconsistent Even After the Temperature is Set to 0}.
\newblock \bibinfo{howpublished}{\url{https://community.openai.com/t/why-the-api-output-is-inconsistent-even-after-the-temperature-is-set-to-0/329541}}.
\newblock
\newblock
\shownote{Accessed: 2024-06-06}.


\bibitem[Dadvar and De~Jong(2012)]%
        {dadvar2012cyberbullying}
\bibfield{author}{\bibinfo{person}{Maral Dadvar} {and} \bibinfo{person}{Franciska De~Jong}.} \bibinfo{year}{2012}\natexlab{}.
\newblock \showarticletitle{Cyberbullying detection: a step toward a safer internet yard}. In \bibinfo{booktitle}{\emph{Proceedings of the 21st International Conference on World Wide Web}}. \bibinfo{pages}{121--126}.
\newblock


\bibitem[Dalianis and Dalianis(2018)]%
        {dalianis2018evaluation}
\bibfield{author}{\bibinfo{person}{Hercules Dalianis} {and} \bibinfo{person}{Hercules Dalianis}.} \bibinfo{year}{2018}\natexlab{}.
\newblock \showarticletitle{Evaluation metrics and evaluation}.
\newblock \bibinfo{journal}{\emph{Clinical Text Mining: secondary use of electronic patient records}} (\bibinfo{year}{2018}), \bibinfo{pages}{45--53}.
\newblock


\bibitem[DeLamar(2012)]%
        {delamar2012supporting}
\bibfield{author}{\bibinfo{person}{Shawna DeLamar}.} \bibinfo{year}{2012}\natexlab{}.
\newblock \bibinfo{booktitle}{\emph{Supporting transition of at-risk students through a freshman orientation model}}.
\newblock \bibinfo{publisher}{Texas A\&M University-Commerce}.
\newblock


\bibitem[Delort et~al\mbox{.}(2011)]%
        {delort2011automatic}
\bibfield{author}{\bibinfo{person}{Jean-Yves Delort}, \bibinfo{person}{Bavani Arunasalam}, {and} \bibinfo{person}{Cecile Paris}.} \bibinfo{year}{2011}\natexlab{}.
\newblock \showarticletitle{Automatic moderation of online discussion sites}.
\newblock \bibinfo{journal}{\emph{International Journal of Electronic Commerce}} \bibinfo{volume}{15}, \bibinfo{number}{3} (\bibinfo{year}{2011}), \bibinfo{pages}{9--30}.
\newblock


\bibitem[D’Alfonso(2020)]%
        {d2020ai}
\bibfield{author}{\bibinfo{person}{Simon D’Alfonso}.} \bibinfo{year}{2020}\natexlab{}.
\newblock \showarticletitle{AI in mental health}.
\newblock \bibinfo{journal}{\emph{Current opinion in psychology}}  \bibinfo{volume}{36} (\bibinfo{year}{2020}), \bibinfo{pages}{112--117}.
\newblock


\bibitem[Faul et~al\mbox{.}(2007a)]%
        {GPower}
\bibfield{author}{\bibinfo{person}{Franz Faul}, \bibinfo{person}{Edgar Erdfelder}, \bibinfo{person}{Axel Buchner}, {and} \bibinfo{person}{Albert-Georg Lang}.} \bibinfo{year}{2007}\natexlab{a}.
\newblock \bibinfo{booktitle}{\emph{GPower 3.1 Manual}}.
\newblock
\urldef\tempurl%
\url{https://www.psychologie.hhu.de/fileadmin/redaktion/Fakultaeten/Mathematisch-Naturwissenschaftliche_Fakultaet/Psychologie/AAP/gpower/GPowerManual.pdf}
\showURL{%
\tempurl}


\bibitem[Faul et~al\mbox{.}(2007b)]%
        {faul2007g}
\bibfield{author}{\bibinfo{person}{Franz Faul}, \bibinfo{person}{Edgar Erdfelder}, \bibinfo{person}{Albert-Georg Lang}, {and} \bibinfo{person}{Axel Buchner}.} \bibinfo{year}{2007}\natexlab{b}.
\newblock \showarticletitle{G* Power 3: A flexible statistical power analysis program for the social, behavioral, and biomedical sciences}.
\newblock \bibinfo{journal}{\emph{Behavior research methods}} \bibinfo{volume}{39}, \bibinfo{number}{2} (\bibinfo{year}{2007}), \bibinfo{pages}{175--191}.
\newblock


\bibitem[Gillespie(2020)]%
        {gillespie2020content}
\bibfield{author}{\bibinfo{person}{Tarleton Gillespie}.} \bibinfo{year}{2020}\natexlab{}.
\newblock \showarticletitle{Content moderation, AI, and the question of scale}.
\newblock \bibinfo{journal}{\emph{Big Data \& Society}} \bibinfo{volume}{7}, \bibinfo{number}{2} (\bibinfo{year}{2020}), \bibinfo{pages}{2053951720943234}.
\newblock


\bibitem[Gopalakrishnan et~al\mbox{.}(2019)]%
        {nonsensitivedata}
\bibfield{author}{\bibinfo{person}{Karthik Gopalakrishnan}, \bibinfo{person}{Behnam Hedayatnia}, \bibinfo{person}{Qinlang Chen}, \bibinfo{person}{Anna Gottardi}, \bibinfo{person}{Sanjeev Kwatra}, \bibinfo{person}{Anu Venkatesh}, \bibinfo{person}{Raefer Gabriel}, {and} \bibinfo{person}{Dilek Hakkani-Tür}.} \bibinfo{year}{2019}\natexlab{}.
\newblock \showarticletitle{{Topical-Chat: Towards Knowledge-Grounded Open-Domain Conversations}}. In \bibinfo{booktitle}{\emph{Proc. Interspeech 2019}}. \bibinfo{pages}{1891--1895}.
\newblock
\urldef\tempurl%
\url{https://doi.org/10.21437/Interspeech.2019-3079}
\showDOI{\tempurl}


\bibitem[Gorwa et~al\mbox{.}(2020)]%
        {gorwa2020algorithmic}
\bibfield{author}{\bibinfo{person}{Robert Gorwa}, \bibinfo{person}{Reuben Binns}, {and} \bibinfo{person}{Christian Katzenbach}.} \bibinfo{year}{2020}\natexlab{}.
\newblock \showarticletitle{Algorithmic content moderation: Technical and political challenges in the automation of platform governance}.
\newblock \bibinfo{journal}{\emph{Big Data \& Society}} \bibinfo{volume}{7}, \bibinfo{number}{1} (\bibinfo{year}{2020}), \bibinfo{pages}{2053951719897945}.
\newblock


\bibitem[He et~al\mbox{.}(2024)]%
        {he2024if}
\bibfield{author}{\bibinfo{person}{Zeyu He}, \bibinfo{person}{Chieh-Yang Huang}, \bibinfo{person}{Chien-Kuang~Cornelia Ding}, \bibinfo{person}{Shaurya Rohatgi}, {and} \bibinfo{person}{Ting-Hao~Kenneth Huang}.} \bibinfo{year}{2024}\natexlab{}.
\newblock \showarticletitle{If in a Crowdsourced Data Annotation Pipeline, a GPT-4}. In \bibinfo{booktitle}{\emph{Proceedings of the CHI Conference on Human Factors in Computing Systems}}. \bibinfo{pages}{1--25}.
\newblock


\bibitem[Hettiachchi and Goncalves(2019)]%
        {hettiachchi2019towards}
\bibfield{author}{\bibinfo{person}{Danula Hettiachchi} {and} \bibinfo{person}{Jorge Goncalves}.} \bibinfo{year}{2019}\natexlab{}.
\newblock \showarticletitle{Towards effective crowd-powered online content moderation}. In \bibinfo{booktitle}{\emph{Proceedings of the 31st Australian Conference on Human-Computer-Interaction}}. \bibinfo{pages}{342--346}.
\newblock


\bibitem[Hickman et~al\mbox{.}(2021)]%
        {hickman2021climate}
\bibfield{author}{\bibinfo{person}{Caroline Hickman}, \bibinfo{person}{Elizabeth Marks}, \bibinfo{person}{Panu Pihkala}, \bibinfo{person}{Susan Clayton}, \bibinfo{person}{R~Eric Lewandowski}, \bibinfo{person}{Elouise~E Mayall}, \bibinfo{person}{Britt Wray}, \bibinfo{person}{Catriona Mellor}, {and} \bibinfo{person}{Lise Van~Susteren}.} \bibinfo{year}{2021}\natexlab{}.
\newblock \showarticletitle{Climate anxiety in children and young people and their beliefs about government responses to climate change: a global survey}.
\newblock \bibinfo{journal}{\emph{The Lancet Planetary Health}} \bibinfo{volume}{5}, \bibinfo{number}{12} (\bibinfo{year}{2021}), \bibinfo{pages}{e863--e873}.
\newblock


\bibitem[Huh et~al\mbox{.}(2013)]%
        {huh2013text}
\bibfield{author}{\bibinfo{person}{Jina Huh}, \bibinfo{person}{Meliha Yetisgen-Yildiz}, {and} \bibinfo{person}{Wanda Pratt}.} \bibinfo{year}{2013}\natexlab{}.
\newblock \showarticletitle{Text classification for assisting moderators in online health communities}.
\newblock \bibinfo{journal}{\emph{Journal of biomedical informatics}} \bibinfo{volume}{46}, \bibinfo{number}{6} (\bibinfo{year}{2013}), \bibinfo{pages}{998--1005}.
\newblock


\bibitem[Jhaver et~al\mbox{.}(2019)]%
        {jhaver2019human}
\bibfield{author}{\bibinfo{person}{Shagun Jhaver}, \bibinfo{person}{Iris Birman}, \bibinfo{person}{Eric Gilbert}, {and} \bibinfo{person}{Amy Bruckman}.} \bibinfo{year}{2019}\natexlab{}.
\newblock \showarticletitle{Human-machine collaboration for content regulation: The case of reddit automoderator}.
\newblock \bibinfo{journal}{\emph{ACM Transactions on Computer-Human Interaction (TOCHI)}} \bibinfo{volume}{26}, \bibinfo{number}{5} (\bibinfo{year}{2019}), \bibinfo{pages}{1--35}.
\newblock


\bibitem[Jozefowicz et~al\mbox{.}(2016)]%
        {jozefowicz2016exploring}
\bibfield{author}{\bibinfo{person}{Rafal Jozefowicz}, \bibinfo{person}{Oriol Vinyals}, \bibinfo{person}{Mike Schuster}, \bibinfo{person}{Noam Shazeer}, {and} \bibinfo{person}{Yonghui Wu}.} \bibinfo{year}{2016}\natexlab{}.
\newblock \showarticletitle{Exploring the limits of language modeling}.
\newblock \bibinfo{journal}{\emph{arXiv preprint arXiv:1602.02410}} (\bibinfo{year}{2016}).
\newblock


\bibitem[Kakuma et~al\mbox{.}(2011)]%
        {kakuma2011human}
\bibfield{author}{\bibinfo{person}{Ritsuko Kakuma}, \bibinfo{person}{Harry Minas}, \bibinfo{person}{Nadja Van~Ginneken}, \bibinfo{person}{Mario~R Dal~Poz}, \bibinfo{person}{Keshav Desiraju}, \bibinfo{person}{Jodi~E Morris}, \bibinfo{person}{Shekhar Saxena}, {and} \bibinfo{person}{Richard~M Scheffler}.} \bibinfo{year}{2011}\natexlab{}.
\newblock \showarticletitle{Human resources for mental health care: current situation and strategies for action}.
\newblock \bibinfo{journal}{\emph{The Lancet}} \bibinfo{volume}{378}, \bibinfo{number}{9803} (\bibinfo{year}{2011}), \bibinfo{pages}{1654--1663}.
\newblock


\bibitem[Kolla et~al\mbox{.}(2024)]%
        {kolla2024llm}
\bibfield{author}{\bibinfo{person}{Mahi Kolla}, \bibinfo{person}{Siddharth Salunkhe}, \bibinfo{person}{Eshwar Chandrasekharan}, {and} \bibinfo{person}{Koustuv Saha}.} \bibinfo{year}{2024}\natexlab{}.
\newblock \showarticletitle{LLM-Mod: Can Large Language Models Assist Content Moderation?}. In \bibinfo{booktitle}{\emph{Extended Abstracts of the CHI Conference on Human Factors in Computing Systems}}. \bibinfo{pages}{1--8}.
\newblock


\bibitem[Lin et~al\mbox{.}(2022)]%
        {lin2022medical}
\bibfield{author}{\bibinfo{person}{Sheryl Lin}, \bibinfo{person}{Albert~C Chong}, \bibinfo{person}{Erin~H Su}, \bibinfo{person}{Sabrina~L Chen}, \bibinfo{person}{Won~Jong Chwa}, \bibinfo{person}{Chantal Young}, \bibinfo{person}{Jacob Schreiber}, {and} \bibinfo{person}{Stephanie~K Zia}.} \bibinfo{year}{2022}\natexlab{}.
\newblock \showarticletitle{Medical student anxiety and depression in the COVID-19 Era: Unique needs of underrepresented students}.
\newblock \bibinfo{journal}{\emph{Education for Health}} \bibinfo{volume}{35}, \bibinfo{number}{2} (\bibinfo{year}{2022}), \bibinfo{pages}{41--47}.
\newblock


\bibitem[Ma et~al\mbox{.}(2023)]%
        {ma2023understanding}
\bibfield{author}{\bibinfo{person}{Zilin Ma}, \bibinfo{person}{Yiyang Mei}, {and} \bibinfo{person}{Zhaoyuan Su}.} \bibinfo{year}{2023}\natexlab{}.
\newblock \showarticletitle{Understanding the benefits and challenges of using large language model-based conversational agents for mental well-being support}. In \bibinfo{booktitle}{\emph{AMIA Annual Symposium Proceedings}}, Vol.~\bibinfo{volume}{2023}. American Medical Informatics Association, \bibinfo{pages}{1105}.
\newblock


\bibitem[Marino et~al\mbox{.}(2016)]%
        {marino2016sharing}
\bibfield{author}{\bibinfo{person}{Casadi Marino}, \bibinfo{person}{Beckie Child}, \bibinfo{person}{Vanessa Campbell~Krasinski}, {et~al\mbox{.}}} \bibinfo{year}{2016}\natexlab{}.
\newblock \showarticletitle{Sharing Experience Learned Firsthand (SELF): Self-disclosure of lived experience in mental health services and supports.}
\newblock \bibinfo{journal}{\emph{Psychiatric Rehabilitation Journal}} \bibinfo{volume}{39}, \bibinfo{number}{2} (\bibinfo{year}{2016}), \bibinfo{pages}{154}.
\newblock


\bibitem[Mart{\'\i}nez(2024)]%
        {martinez2024re}
\bibfield{author}{\bibinfo{person}{Eric Mart{\'\i}nez}.} \bibinfo{year}{2024}\natexlab{}.
\newblock \showarticletitle{Re-evaluating GPT-4’s bar exam performance}.
\newblock \bibinfo{journal}{\emph{Artificial Intelligence and Law}} (\bibinfo{year}{2024}), \bibinfo{pages}{1--24}.
\newblock


\bibitem[Milne et~al\mbox{.}(2019)]%
        {milne2019improving}
\bibfield{author}{\bibinfo{person}{David~N Milne}, \bibinfo{person}{Kathryn~L McCabe}, {and} \bibinfo{person}{Rafael~A Calvo}.} \bibinfo{year}{2019}\natexlab{}.
\newblock \showarticletitle{Improving moderator responsiveness in online peer support through automated triage}.
\newblock \bibinfo{journal}{\emph{Journal of medical Internet research}} \bibinfo{volume}{21}, \bibinfo{number}{4} (\bibinfo{year}{2019}), \bibinfo{pages}{e11410}.
\newblock


\bibitem[Milne et~al\mbox{.}(2016)]%
        {milne2016clpsych}
\bibfield{author}{\bibinfo{person}{David~N Milne}, \bibinfo{person}{Glen Pink}, \bibinfo{person}{Ben Hachey}, {and} \bibinfo{person}{Rafael~A Calvo}.} \bibinfo{year}{2016}\natexlab{}.
\newblock \showarticletitle{Clpsych 2016 shared task: Triaging content in online peer-support forums}. In \bibinfo{booktitle}{\emph{Proceedings of the third workshop on computational linguistics and clinical psychology}}. \bibinfo{pages}{118--127}.
\newblock


\bibitem[Musicant et~al\mbox{.}(2003)]%
        {musicant2003optimizing}
\bibfield{author}{\bibinfo{person}{David~R Musicant}, \bibinfo{person}{Vipin Kumar}, \bibinfo{person}{Aysel Ozgur}, {et~al\mbox{.}}} \bibinfo{year}{2003}\natexlab{}.
\newblock \showarticletitle{Optimizing F-Measure with Support Vector Machines.}. In \bibinfo{booktitle}{\emph{FLAIRS}}. \bibinfo{pages}{356--360}.
\newblock


\bibitem[Nahmias and Perel(2021)]%
        {nahmias2021oversight}
\bibfield{author}{\bibinfo{person}{Yifat Nahmias} {and} \bibinfo{person}{Maayan Perel}.} \bibinfo{year}{2021}\natexlab{}.
\newblock \showarticletitle{The oversight of content moderation by AI: Impact assessments and their limitations}.
\newblock \bibinfo{journal}{\emph{Harv. J. on Legis.}}  \bibinfo{volume}{58} (\bibinfo{year}{2021}), \bibinfo{pages}{145}.
\newblock


\bibitem[Naqshbandi et~al\mbox{.}(2016)]%
        {naqshbandi2016helping}
\bibfield{author}{\bibinfo{person}{Khushnood Naqshbandi}, \bibinfo{person}{David~N Milne}, \bibinfo{person}{Ben Davies}, \bibinfo{person}{Sophie Potter}, \bibinfo{person}{Rafael~A Calvo}, {and} \bibinfo{person}{Simon Hoermann}.} \bibinfo{year}{2016}\natexlab{}.
\newblock \showarticletitle{Helping young people going through tough times: Perspectives for a peer-to-peer chat support system}. In \bibinfo{booktitle}{\emph{Proceedings of the 28th Australian conference on computer-human interaction}}. \bibinfo{pages}{640--642}.
\newblock


\bibitem[O'dea et~al\mbox{.}(2015)]%
        {o2015detecting}
\bibfield{author}{\bibinfo{person}{Bridianne O'dea}, \bibinfo{person}{Stephen Wan}, \bibinfo{person}{Philip~J Batterham}, \bibinfo{person}{Alison~L Calear}, \bibinfo{person}{Cecile Paris}, {and} \bibinfo{person}{Helen Christensen}.} \bibinfo{year}{2015}\natexlab{}.
\newblock \showarticletitle{Detecting suicidality on Twitter}.
\newblock \bibinfo{journal}{\emph{Internet Interventions}} \bibinfo{volume}{2}, \bibinfo{number}{2} (\bibinfo{year}{2015}), \bibinfo{pages}{183--188}.
\newblock


\bibitem[OpenAI(2023)]%
        {openai_2023_ft}
\bibfield{author}{\bibinfo{person}{OpenAI}.} \bibinfo{year}{2023}\natexlab{}.
\newblock \bibinfo{title}{Preparing Your Dataset}.
\newblock \bibinfo{howpublished}{\url{https://platform.openai.com/docs/guides/fine-tuning/preparing-your-dataset}}.
\newblock
\newblock
\shownote{Accessed: 2024-06-25}.


\bibitem[{OpenAI}(2024)]%
        {openai_privacy_policy}
\bibfield{author}{\bibinfo{person}{{OpenAI}}.} \bibinfo{year}{2024}\natexlab{}.
\newblock \bibinfo{title}{OpenAI Privacy Policy}.
\newblock \bibinfo{howpublished}{\url{https://openai.com/policies/privacy-policy/}}.
\newblock
\newblock
\shownote{Accessed: 2024-06-13}.


\bibitem[Organization et~al\mbox{.}(2017)]%
        {world2017depression}
\bibfield{author}{\bibinfo{person}{World~Health Organization} {et~al\mbox{.}}} \bibinfo{year}{2017}\natexlab{}.
\newblock \bibinfo{booktitle}{\emph{Depression and other common mental disorders: global health estimates}}.
\newblock \bibinfo{type}{{T}echnical {R}eport}. \bibinfo{institution}{World Health Organization}.
\newblock


\bibitem[Patel et~al\mbox{.}(2007)]%
        {patel2007mental}
\bibfield{author}{\bibinfo{person}{Vikram Patel}, \bibinfo{person}{Alan~J Flisher}, \bibinfo{person}{Sarah Hetrick}, {and} \bibinfo{person}{Patrick McGorry}.} \bibinfo{year}{2007}\natexlab{}.
\newblock \showarticletitle{Mental health of young people: a global public-health challenge}.
\newblock \bibinfo{journal}{\emph{The lancet}} \bibinfo{volume}{369}, \bibinfo{number}{9569} (\bibinfo{year}{2007}), \bibinfo{pages}{1302--1313}.
\newblock


\bibitem[Siapera(2022)]%
        {siapera2022ai}
\bibfield{author}{\bibinfo{person}{Eugenia Siapera}.} \bibinfo{year}{2022}\natexlab{}.
\newblock \showarticletitle{AI content moderation, racism and (de) coloniality}.
\newblock \bibinfo{journal}{\emph{International journal of bullying prevention}} \bibinfo{volume}{4}, \bibinfo{number}{1} (\bibinfo{year}{2022}), \bibinfo{pages}{55--65}.
\newblock


\bibitem[Siegert et~al\mbox{.}(2023)]%
        {siegert2023depression}
\bibfield{author}{\bibinfo{person}{Richard~J Siegert}, \bibinfo{person}{Ajit Narayanan}, \bibinfo{person}{Joanna Dipnall}, \bibinfo{person}{Lisa Gossage}, \bibinfo{person}{Wendy Wrapson}, \bibinfo{person}{Alexander Sumich}, \bibinfo{person}{Fabrice Merien}, \bibinfo{person}{Michael Berk}, \bibinfo{person}{Janis Paterson}, {and} \bibinfo{person}{El-Shadan Tautolo}.} \bibinfo{year}{2023}\natexlab{}.
\newblock \showarticletitle{Depression, anxiety and worry in young Pacific adults in New Zealand during the COVID-19 pandemic}.
\newblock \bibinfo{journal}{\emph{Australian \& New Zealand Journal of Psychiatry}} \bibinfo{volume}{57}, \bibinfo{number}{5} (\bibinfo{year}{2023}), \bibinfo{pages}{698--709}.
\newblock


\bibitem[Slonje et~al\mbox{.}(2013)]%
        {slonje2013nature}
\bibfield{author}{\bibinfo{person}{Robert Slonje}, \bibinfo{person}{Peter~K Smith}, {and} \bibinfo{person}{Ann Fris{\'e}n}.} \bibinfo{year}{2013}\natexlab{}.
\newblock \showarticletitle{The nature of cyberbullying, and strategies for prevention}.
\newblock \bibinfo{journal}{\emph{Computers in human behavior}} \bibinfo{volume}{29}, \bibinfo{number}{1} (\bibinfo{year}{2013}), \bibinfo{pages}{26--32}.
\newblock


\bibitem[Thompson et~al\mbox{.}(2021)]%
        {transition}
\bibfield{author}{\bibinfo{person}{Miles Thompson}, \bibinfo{person}{Chris Pawson}, {and} \bibinfo{person}{Bethan Evans}.} \bibinfo{year}{2021}\natexlab{}.
\newblock \showarticletitle{Navigating entry into higher education: the transition to independent learning and living}.
\newblock \bibinfo{journal}{\emph{Journal of Further and Higher Education}} \bibinfo{volume}{45}, \bibinfo{number}{10} (\bibinfo{year}{2021}), \bibinfo{pages}{1398--1410}.
\newblock


\bibitem[Udupa et~al\mbox{.}(2023)]%
        {udupa2023ethical}
\bibfield{author}{\bibinfo{person}{Sahana Udupa}, \bibinfo{person}{Antonis Maronikolakis}, {and} \bibinfo{person}{Axel Wisiorek}.} \bibinfo{year}{2023}\natexlab{}.
\newblock \showarticletitle{Ethical scaling for content moderation: Extreme speech and the (in) significance of artificial intelligence}.
\newblock \bibinfo{journal}{\emph{Big Data \& Society}} \bibinfo{volume}{10}, \bibinfo{number}{1} (\bibinfo{year}{2023}), \bibinfo{pages}{20539517231172424}.
\newblock


\bibitem[Vacchiano(2023)]%
        {vacchiano2023first}
\bibfield{author}{\bibinfo{person}{Mattia Vacchiano}.} \bibinfo{year}{2023}\natexlab{}.
\newblock \showarticletitle{How the first COVID-19 lockdown worsened younger generations’ mental health: Insights from network theory}.
\newblock \bibinfo{journal}{\emph{Sociological Research Online}} \bibinfo{volume}{28}, \bibinfo{number}{3} (\bibinfo{year}{2023}), \bibinfo{pages}{884--893}.
\newblock


\bibitem[VM et~al\mbox{.}(2024)]%
        {vm2024fine}
\bibfield{author}{\bibinfo{person}{Kushala VM}, \bibinfo{person}{Harikrishna Warrier}, \bibinfo{person}{Yogesh Gupta}, {et~al\mbox{.}}} \bibinfo{year}{2024}\natexlab{}.
\newblock \showarticletitle{Fine Tuning LLM for Enterprise: Practical Guidelines and Recommendations}.
\newblock \bibinfo{journal}{\emph{arXiv preprint arXiv:2404.10779}} (\bibinfo{year}{2024}).
\newblock


\bibitem[Wang et~al\mbox{.}(2023)]%
        {wang2023not}
\bibfield{author}{\bibinfo{person}{Yuxia Wang}, \bibinfo{person}{Haonan Li}, \bibinfo{person}{Xudong Han}, \bibinfo{person}{Preslav Nakov}, {and} \bibinfo{person}{Timothy Baldwin}.} \bibinfo{year}{2023}\natexlab{}.
\newblock \showarticletitle{Do-not-answer: A dataset for evaluating safeguards in llms}.
\newblock \bibinfo{journal}{\emph{arXiv preprint arXiv:2308.13387}} (\bibinfo{year}{2023}).
\newblock


\bibitem[{WoeBot Health Ltd.}(5 03)]%
        {WoeBotWebsite}
\bibfield{author}{\bibinfo{person}{{WoeBot Health Ltd.}}} \bibinfo{year}{Accessed: 2024-05-03}\natexlab{}.
\newblock \bibinfo{title}{{WoeBot}}.
\newblock \bibinfo{howpublished}{\url{https://woebothealth.com/}}.
\newblock


\bibitem[Wu et~al\mbox{.}(2023)]%
        {wu2023brief}
\bibfield{author}{\bibinfo{person}{Tianyu Wu}, \bibinfo{person}{Shizhu He}, \bibinfo{person}{Jingping Liu}, \bibinfo{person}{Siqi Sun}, \bibinfo{person}{Kang Liu}, \bibinfo{person}{Qing-Long Han}, {and} \bibinfo{person}{Yang Tang}.} \bibinfo{year}{2023}\natexlab{}.
\newblock \showarticletitle{A brief overview of ChatGPT: The history, status quo and potential future development}.
\newblock \bibinfo{journal}{\emph{IEEE/CAA Journal of Automatica Sinica}} \bibinfo{volume}{10}, \bibinfo{number}{5} (\bibinfo{year}{2023}), \bibinfo{pages}{1122--1136}.
\newblock


\bibitem[{Wysa}(5 03)]%
        {WysaWebsite}
\bibfield{author}{\bibinfo{person}{{Wysa}}.} \bibinfo{year}{Accessed: 2024-05-03}\natexlab{}.
\newblock \bibinfo{title}{{Wysa}}.
\newblock \bibinfo{howpublished}{\url{https://www.wysa.com/}}.
\newblock


\bibitem[Zhao et~al\mbox{.}(2016)]%
        {zhao2016automatic}
\bibfield{author}{\bibinfo{person}{Rui Zhao}, \bibinfo{person}{Anna Zhou}, {and} \bibinfo{person}{Kezhi Mao}.} \bibinfo{year}{2016}\natexlab{}.
\newblock \showarticletitle{Automatic detection of cyberbullying on social networks based on bullying features}. In \bibinfo{booktitle}{\emph{Proceedings of the 17th international conference on distributed computing and networking}}. \bibinfo{pages}{1--6}.
\newblock


\bibitem[Zhu et~al\mbox{.}(2024)]%
        {zhu2024can}
\bibfield{author}{\bibinfo{person}{Yilun Zhu}, \bibinfo{person}{Joel Ruben~Antony Moniz}, \bibinfo{person}{Shruti Bhargava}, \bibinfo{person}{Jiarui Lu}, \bibinfo{person}{Dhivya Piraviperumal}, \bibinfo{person}{Site Li}, \bibinfo{person}{Yuan Zhang}, \bibinfo{person}{Hong Yu}, {and} \bibinfo{person}{Bo-Hsiang Tseng}.} \bibinfo{year}{2024}\natexlab{}.
\newblock \showarticletitle{Can Large Language Models Understand Context?}
\newblock \bibinfo{journal}{\emph{arXiv preprint arXiv:2402.00858}} (\bibinfo{year}{2024}).
\newblock


\end{thebibliography}
